\documentclass[11pt]{article}
\usepackage[left=1in,top=1in,right=1in,bottom=1in,letterpaper]{geometry}
\usepackage[colorlinks,linkcolor=red,filecolor=blue,citecolor=blue,urlcolor=blue]{hyperref}
\usepackage{mathtools}
\usepackage{url}
\usepackage{amsthm,amsmath,amssymb,amscd}
\usepackage{mathrsfs}
\usepackage{amsfonts}
\usepackage[utf8]{inputenc}
\usepackage{graphicx}
\usepackage{float}
\usepackage{bbding}
\everymath{\displaystyle}
\usepackage{parskip}
\usepackage{subcaption}
\usepackage{enumerate}
\usepackage{bm,bbm}
\usepackage[shortlabels]{enumitem}
\usepackage{algorithm,algorithmic}
\usepackage{dsfont}
\usepackage{booktabs}
\usepackage{array,siunitx}
\usepackage{xspace}
\usepackage{pifont}
\usepackage{xcolor}
\usepackage{colortbl}
\usepackage{aliascnt}
\usepackage{cleveref}
\usepackage{mathpazo}
\usepackage{natbib}
\numberwithin{equation}{section}
\usepackage{apptools}
\AtAppendix{\counterwithin{thm}{section}}

\usepackage{amsmath,amsfonts,bm}

\def\1{\bm{1}}

\def\vzero{{\bm{0}}}
\def\vone{{\bm{1}}}

\def\vtheta{{\bm{\theta}}}

\def\vb{{\bm{b}}}

\def\ve{{\bm{e}}}

\def\vh{{\bm{h}}}

\def\vp{{\bm{p}}}

\def\vr{{\bm{r}}}

\def\vu{{\bm{u}}}
\def\vv{{\bm{v}}}
\def\vw{{\bm{w}}}
\def\vx{{\bm{x}}}
\def\vy{{\bm{y}}}
\def\vz{{\bm{z}}}
\def\vbeta{{\bm{\beta}}}
\def\vpsi{{\bm{\psi}}}

\def\vDelta{{\bm{\Delta}}}

\def\vphi{{\bm{\phi}}}
\def\vX{{\bm{X}}}
\def\vZ{{\bm{Z}}}
\def\vW{{\bm{W}}}

\def\vE{{\bm{E}}}
\def\vL{{\bm{L}}}

\def\mA{{\bm{A}}}

\def\mC{{\bm{C}}}

\def\mE{{\bm{E}}}

\def\mG{{\bm{G}}}
\def\mH{{\bm{H}}}
\def\mI{{\bm{I}}}

\def\mK{{\bm{K}}}
\def\mL{{\bm{L}}}

\def\mQ{{\bm{Q}}}

\def\mV{{\bm{V}}}
\def\mW{{\bm{W}}}

\def\msigma{{\bm{\sigma}}}
\def\mPsi{{\bm{\Psi}}}

\DeclareMathAlphabet{\mathsfit}{\encodingdefault}{\sfdefault}{m}{sl}
\SetMathAlphabet{\mathsfit}{bold}{\encodingdefault}{\sfdefault}{bx}{n}

\newcommand{\E}{\mathbb{E}}

\newcommand{\R}{\mathbb{R}}

\newcommand{\dif}{\mathrm{d}}

\newtheorem{thm}{Theorem}[section]

\newtheorem{Assumption}{Assumption}
\newaliascnt{lemma}{thm}
\newtheorem{lemma}[lemma]{Lemma}
\aliascntresetthe{lemma}
\newaliascnt{coro}{thm}

\aliascntresetthe{coro}
\newaliascnt{prop}{thm}
\newtheorem{prop}[prop]{Proposition}
\aliascntresetthe{prop}
\theoremstyle{definition}
\newtheorem{definition}{Definition}[section]

\crefname{equation}{eq.}{eqs.}
\Crefname{equation}{Equation}{Equations}
\crefname{Assumption}{Assumption}{Assumptions}
\Crefname{Assumption}{Assumption}{Assumptions}
\crefname{thm}{Theorem}{Theorems}
\Crefname{thm}{Theorem}{Theorems}
\crefname{lemma}{Lemma}{Lemmas}
\Crefname{lemma}{Lemma}{Lemmas}
\crefname{coro}{Corollary}{Corollaries}
\Crefname{coro}{Corollary}{Corollaries}
\crefname{prop}{Proposition}{Propositions}
\Crefname{prop}{Proposition}{Propositions}
\crefname{definition}{def.}{defs.}
\Crefname{definition}{Def.}{Defs.}
\crefname{algorithm}{alg.}{algs.}
\Crefname{algorithm}{Alg.}{Algs.}
\crefname{figure}{Figure}{Figures}
\Crefname{figure}{Figure}{Figures}
\crefname{table}{Table}{Tables}
\crefname{section}{Section}{Sections}
\Crefname{section}{Section}{Sections}
\crefname{appendix}{Appendix}{Appendices}
\Crefname{appendix}{Appendix}{Appendices}
\crefname{subappendix}{Appendix}{Appendices}
\Crefname{subappendix}{Appendix}{Appendices}

\renewcommand{\P}{\mathbb{P}}
\usepackage{colortbl}
\definecolor{Ocean}{RGB}{129,194,234}
\newcommand{\Sig}{\text{Sig}}
\newcommand{\TF}{\text{TF}}
\newcommand{\Attn}{\text{Attn}}
\newcommand{\MLP}{\text{MLP}}
\newcommand{\norm}[1]{\lVert #1\rVert}
\newcommand{\abs}[1]{\lvert #1\rvert}
\newcommand{\supp}{\operatorname{supp}}
\newcommand{\TV}{\mathrm{TV}}
\newcommand{\op}{\mathrm{op}}
\newcommand{\soft}{\mathsf S}
\newcommand{\detval}[1]{#1_{\phantom{\pm 0.000}}}

\title{Transformers with Signatures: Towards Universal In-Context Learning in Path Space}

\author{
Ziheng Cheng \thanks{University of California, Berkeley, Department of IEOR. Email: \texttt{ziheng\_cheng@berkeley.edu}} 
\and 
Xin Guo \thanks{University of California, Berkeley, Department of IEOR. Email: \texttt{xinguo@berkeley.edu}}
\and
Ben Hambly \thanks{University of Oxford,
Mathematical Institute. Email: \texttt{ben.hambly@maths.ox.ac.uk}}
}
\date{}

\begin{document}

\maketitle

\begin{abstract}
In-context learning (ICL) enables transformers to adapt to new tasks from demonstrations without updating their parameters. However, theoretical guarantees for transformers remain limited in general,  and elusive in particular when target responses depend nonlinearly on the evolution of observed paths. 
We establish an ICL framework for path space, using path signature features.
Building on the universal linear approximation property of signatures, we construct explicitly a causal softmax transformer with two attention heads per layer that emulates in-context Lasso. Under sparse signature approximation and suitable moment and concentration conditions, we establish finite-sample prediction error bounds.
Our analysis goes beyond the IID data setting in the existing ICL theory, allowing dependence among demonstrations and the query. 

Experiments on in-context sparse recovery, nonlinear path regression, execution-cost prediction, and financial payoff inference demonstrate robustness to changes in observation grids and adaptation to unseen market
environments. 
Across these prediction tasks, transformers with signatures
match or outperform vanilla transformer baselines with substantially fewer trainable parameters, highlighting the effectiveness and  efficiency of signature-based ICL.
\end{abstract}

\section{Introduction}

\paragraph{In-context learning with transformers.}
In-context learning (ICL) is an AI paradigm in which a large language model (LLM) learns to perform a new task dynamically by looking at a few input-output examples, without changing its internal model weights, i.e., without retraining its parameters \citep{brown2020language, garg2022what,akyurek2023learning}.
The transformer architecture is the primary engine that enables ICL in modern LLMs. A transformer processes a sequence of vectors called tokens,  updates their vector representations through successive layers, and extracts prediction after the last layer. For example, a linear regression prompt in ICL contains several input--label pairs from an unknown linear function, followed
by a new query token. The goal is to predict the corresponding response. 

 Pre-trained transformers can solve unseen regression tasks in context, matching the sample efficiency of canonical estimators such as ordinary least squares and the Lasso for linear models \citep{garg2022what, akyurek2023learning}.
 {Theoretically}, a growing body of constructive work explains this capability by showing that transformers can explicitly emulate iterative optimization algorithms with appropriate parameter configurations and attention layers  
\citep{akyurek2023learning, vonoswald2023transformers, bai2023transformers}.

However, existing work on ICL theory leaves two critical gaps. \emph{From a theoretical perspective}, most constructive analyses focus strictly on linear regression. 
Recent 
extensions to nonlinear functions via neural feature maps \citep{kim2024minimax,li2025transformers}
 rely on independently sampled static inputs and provide no prediction
guarantees for nonlinear functionals of stochastic paths with 
temporal dependence. In practice, however, the demonstrations may consist of successive segments of a common stochastic process. The learning problem must therefore account both for nonlinear dependence on each observed path and for statistical dependence among the demonstration paths and the query.
\emph{From an interpretability perspective}, pure approximation guarantees do
not, by themselves, explain what the features represent.
The black-box neural features in these constructions 
do not have an intrinsic interpretation in terms of the observed path.

The question is: can a transformer with fixed parameters learn nonlinear functionals of stochastic paths in context,  under statistical dependence across demonstrations and the query?

\paragraph{Path signatures and financial applications.}
Originally introduced and studied in algebraic topology \citep{chen1954iterated,Chen1957}, the path signature, also known as the signature transform or signature \citep{lyons2007differential,friz2010multidimensional}, 
has proven to be a powerful tool for feature generation and pattern recognition with state-of-the-art performance  in machine learning. 
(See for instance, \citep{bonnier2019deep, chevyrev2022signature, chevyrev2025primer,lyons2022signature, gu2024transportation, guo2025consistency} and the references therein).

The signature has recently been widely adopted in finance, for example, for information extraction from financial time series data \citep{gyurko2014extracting},
for the calibration of signature-based asset-price models \citep{cuchiero2023signature,cuchiero2025universal,ceylan2026universal}, for derivative  pricing and hedging under signature volatility models \citep{perezarribas2018derivatives, abijaber2025signature}, in signature-kernel methods for path-dependent hedging \citep{cirone2025rough, guo2026tradable}, and for optimal execution \citep{kalsi2020optimal}.

Many of the empirical successes, as well as theoretical studies, are built upon the signature's striking  {\em universal nonlinearity} property \citep{lyons2014rough}: any continuous function of the time series can be approximated arbitrarily well by a linear combination of its signature components.
This property linearizes the problem of feature selection and does not require the engineering of neural network architectures for  nonlinear and path-dependent functions, with the added benefit of interpretability \citep{cuchiero2025universal,ceylan2026universal}.
Most recently, this property has been shown to be efficient with path-dependent objective functions in \citep{guo2026signature} for contextual bandit problems  and in \citep{guo2026tradable} for hedging problems.

\paragraph{Integrating ICL and path signatures.}

For many practical problems,
it is natural to combine the universal nonlinearity property of the signature with ICL. 
On one hand, signature features provide an \emph{expressive representation of path geometry}. 
For instance, lead-lag signatures may encode quadratic variation through second-order terms. 
Meanwhile, ICL complements this representation with a mechanism for \emph{adaptation to unseen stochastic environments}, using demonstrations to infer the relationship prevailing under new conditions. 
Moreover, because signature coordinates are defined as path integrals, evaluating the same feature map across varying sampling frequencies and irregular timestamps provides natural \emph{robustness to observation grids}.

The universal nonlinearity of the signature, the demonstrated efficiency of the signature for path-dependent objectives, and the adaptive power of ICL, motivate us to focus on the following question. 
Can a transformer explicitly implement in-context Lasso on a shared signature representation with a prediction error bound?

\paragraph{Our work.}
In this paper, we explore the integration of signatures and ICL for transformers.

\begin{itemize}
\item  We construct a causal softmax transformer that uses the signature as path embedding and simulates in-context Lasso. The transformer is explicit and simple: it consists of two softmax heads per layer and an MLP that performs soft-thresholding and projection exactly. 

This construction is novel. All prior theoretical analyses of in-context Lasso rely predominantly on unmasked attention \citep{hu2025softmax}, linear attention \citep{vonoswald2023transformers}, or ReLU or Sigmoid attention \citep{bai2023transformers,li2025transformers}, whereas masked softmax attention is the standard mechanism used in frontier LLMs.
 
\item We establish a universal in-context prediction guarantee for general nonlinear, path-dependent
target functions via signature features (Theorem \ref{thm:sm-main}).
More precisely, with $N$ in-context demonstrations the mean square error can be bounded by $
    \mathcal{O}(\epsilon^2 + 1/N + \epsilon_{\rm TF}^2)$,
consisting of the signature universal approximation error $\epsilon^2$, the Lasso statistical estimation rate $1/N$, 
and an arbitrarily small transformer approximation error $\epsilon_{\rm TF}^2$.

Our prediction guarantee 
goes beyond the IID data in the existing ICL theory, 
allowing dependence among the demonstrations and the query.
Even within the high-dimensional Lasso literature
\citep{han2020dependent,adamek2023lasso}, statistical analyses of dependent data typically
rely on indirect assumptions governing decaying temporal correlations. 
In contrast, we formulate direct moment and concentration conditions on the signature features and verify them for dissipative diffusions, the CIR process, and fractional OU processes (\cref{prop:sm-brownian-sde,prop:sm-cir-concentration,prop:sm-fou-concentration}).

\item Our experiments confirm that transformers with signatures can simulate in-context Lasso for sparse feature recovery and perform path-dependent ICL  effectively with a compact architecture. Across extensive financial tasks, the prediction accuracy is comparable to or better than that of vanilla transformers, while requiring substantially fewer trainable parameters. 
We also show that signature-based ICL is robust with respect to changes in the path observation grids.

\end{itemize}


Overall, the signature transform enhances the transformer's capability by reducing the burden of learning a path representation from scratch.
Meanwhile, the transformer provides essential expressivity when the signature truncation level is insufficient to fully linearize the target functional, while ICL leverages cross-task priors from pre-training beyond single-task signature Lasso fitting.

\section{Preliminaries}
\label{sec:preliminaries}


\subsection{Transformer Architecture}
A transformer updates token representations through successive layers, each comprising a causal multi-head attention block followed by a multilayer perceptron (MLP). The attention block combines information across sequence positions, while the MLP applies a nonlinear transformation to each token representation. The predictions are then computed from the final layer’s outputs.
Rather than computing a single attention map over the full embedding space, multi-head attention operates several attention heads in parallel to capture distinct relational patterns across the sequence. 
Specifically, each head projects the input token embeddings into lower-dimensional subspaces via separate learnable weight matrices for queries ($Q$), keys ($K$) and values ($V$).

We write a sequence of $N$ tokens of dimension $D$ as
$\mH=[\vh_0,\ldots,\vh_{N-1}]^\top\in\R^{N\times D}$.

\begin{figure}[t]
    \centering
    \includegraphics[width=\textwidth]{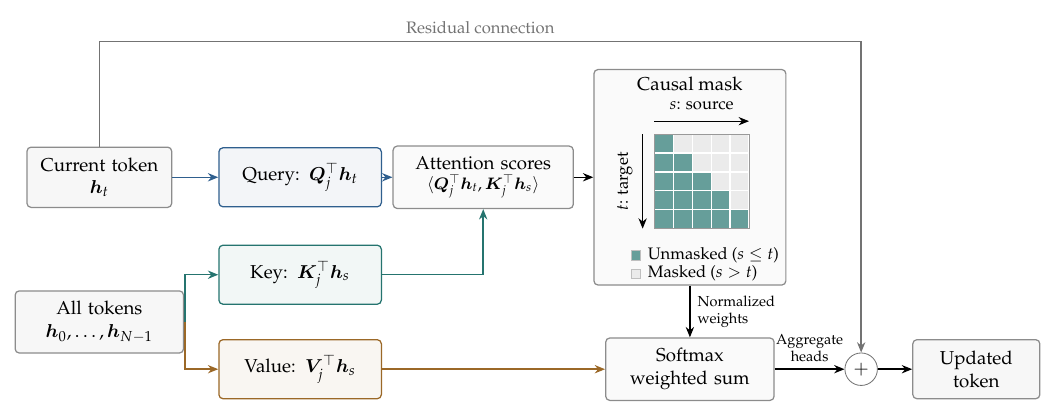}
    \caption{Attention computation at position $t$. The query and key vectors determine the attention scores through their inner products. The causal mask restricts each token to information from itself and preceding tokens. Each head forms a weighted average of the corresponding value vectors; the head outputs are summed and
    added to $\vh_t$, as in
    \cref{eq:masked-softmax-attention-vector}.}
    \label{fig:attention-computation}
\end{figure}

\begin{definition}[Masked Softmax Attention]\label{def:masked-softmax-attention}
A masked softmax attention block $\Attn_\vtheta(\cdot)$ with $M$ heads and head dimension $d_h$ has parameters
$\vtheta=\{(\mQ_j,\mK_j,\mV_j)\}_{j=1}^M$, where
$\mQ_j,\mK_j\in\R^{D\times d_h}$ and
$\mV_j\in\R^{D\times D}$. On any input sequence $\mH=[\vh_0,\ldots,\vh_{N-1}]^\top\in\R^{N\times D}$,
\begin{equation}\label{eq:masked-softmax-attention}
    \Attn_\vtheta(\mH)
    =\mH+\sum_{j=1}^M
      \operatorname{softmax}
      \!\big(\operatorname{Mask}\big(\mH\mQ_j(\mH\mK_j)^\top\big)\big)
      \mH\mV_j\in\R^{N\times D}.
\end{equation}
Here $\operatorname{Mask}(\cdot)$ stands for the causal mask operator: for a matrix $\mA=(A_{ts})\in\R^{N\times N}$, 
\begin{equation}
    [\operatorname{Mask}(\mA)]_{ts}:=\begin{cases}A_{ts}, & s\leq t,\\ -\infty, & s>t.\end{cases}
\end{equation}
Equivalently, for $0\leq t<N$,
\begin{equation}\label{eq:masked-softmax-attention-vector}
    [\Attn_\vtheta(\mH)]_t
    =\vh_t+\sum_{j=1}^M\sum_{s=0}^t
       \frac{\exp\!\big(\langle\mQ_j^\top\vh_t,\mK_j^\top\vh_s\rangle\big)}
    {\sum_{r=0}^t
      \exp\!\big(\langle\mQ_j^\top\vh_t,\mK_j^\top\vh_r\rangle\big)}\mV_j^\top\vh_s.
\end{equation}
\end{definition}
\Cref{fig:attention-computation} illustrates the procedures of attention computation.
Each attention head uses three linear projections: query, key, and
value. At position $t$ head $j$ matches the query
$\mQ_j^\top\vh_t$ with the keys $\mK_j^\top\vh_s$. The inner product
$\langle\mQ_j^\top\vh_t,\mK_j^\top\vh_s\rangle$ is the attention score
for position $s$, while $\mV_j^\top\vh_s$ is the corresponding value vector. Queries and keys therefore determine how much weight a token $\vh_s$ receives, while values determine the information
it contributes. 
Each head has its own projections $(\mQ_j,\mK_j,\mV_j)$, allowing the heads to combine information in different ways. Their outputs lie in $\R^D$ and are summed directly. 

Importantly, the causal mask restricts the information available at position $t$ to
the first $t+1$ tokens in the prompt, $\vh_0,\ldots,\vh_t$.
A query placed after the demonstrations can, therefore, use information from all of them, while earlier tokens cannot access the query.

It is worth reiterating that prior theoretical ICL constructions focus on unmasked, or linear attention,  or ReLU, or Sigmoid attention, whereas our analysis will be for masked softmax attention used in SOTA LLMs.

After the attention block, the MLP block further processes each individual token, without mixing information between tokens. Here we use the activation $\operatorname{ReLU}(a)=\max\{a,0\}$.
\begin{definition}[MLP]\label{def:mlp}
An MLP block with hidden width $D'$, denoted by $\MLP_\vtheta(\cdot)$, has parameters
$\vtheta=(\mW_1,\mW_2)$, where
$\mW_1,\mW_2\in\R^{D\times D'}$. For any input sequence $\mH=[\vh_0,\ldots,\vh_{N-1}]^\top\in\R^{N\times D}$,
\begin{equation}
    \MLP_\vtheta(\mH)=\mH+\operatorname{ReLU}(\mH\mW_1)\mW_2^\top.
\end{equation}
We can write 
in vector form  
$[\MLP_\vtheta(\mH)]_t=\vh_t+\mW_2\operatorname{ReLU}(\mW_1^\top \vh_t)$.
\end{definition}

\begin{definition}[Transformer]\label{def:transformer}
    An $L$-layer transformer with parameters
    $\vtheta=\{\vtheta_{\rm attn}^{(\ell)},\vtheta_{\rm mlp}^{(\ell)}\}_{\ell=1}^L$
    maps $\mH^{(0)}\in\R^{N\times D}$ to
    $\mH^{(L)}=\TF_\vtheta(\mH^{(0)})\in\R^{N\times D}$ through
    \begin{equation}
        \mH^{(\ell)}=\MLP_{\vtheta_{\rm mlp}^{(\ell)}}\Big(
        \Attn_{\vtheta_{\rm attn}^{(\ell)}}(\mH^{(\ell-1)})\Big),
        \qquad \ell=1,\ldots,L.
    \end{equation}
\end{definition}
The MLP acts individually on each token, so the causal restriction from masked attention is preserved throughout all $L$ layers.

The attention and MLP blocks play complementary roles in transformers.
Attention retrieves and aggregates information from the current and
preceding tokens, while MLP further processes the resulting representation through nonlinearity.

\subsection{Path Representation by the Signature}

Let $\vx:[0,1]\to\R^d$ be a continuous path with
$\vx_0=\vzero$.\footnote{Otherwise, one can prepend the origin
$(0,\vzero)$ to the time-augmented path \citep{chevyrev2025primer}.}
The time-augmented path is
\begin{equation}\label{eq:sm-path-augmentation}
    \widetilde{\vx}_t:=(t,\vx_t),\qquad 0\leq t\leq1.
\end{equation}
\begin{definition}[Signature transform]
\label{def:sm-signature}
The signature of the time-augmented path is the sequence
$\Sig(\widetilde{\vx})=(\Sig^{(k)}(\widetilde{\vx}))_{k\geq0}$.
The zeroth level is $\Sig^{(0)}(\widetilde{\vx}):=1$.
For $k\geq1$, the $k$-th level $\Sig^{(k)}(\widetilde{\vx})\in(\R^{d+1})^{\otimes k}$ is defined coordinatewise by
\begin{equation}\label{eq:sm-signature-levels}
    \bigl[\Sig^{(k)}(\widetilde{\vx})\bigr]_{i_1,\ldots,i_k}:=
    \int_{0<u_1<\cdots<u_k<1}
       \dif \widetilde{x}_{u_1}^{(i_1)}\cdots\dif \widetilde{x}_{u_k}^{(i_k)},
    \qquad i_1,\ldots,i_k\in\{0,\ldots,d\}.
\end{equation}
For $M\geq0$, the truncated signature transform of $\widetilde{\vx}$ up to level $M$ is defined as
\begin{equation}\label{eq:sm-full-signature}
    \Sig_M(\widetilde{\vx}):=\Big(1,\Sig^{(1)}(\widetilde{\vx}),\ldots,\Sig^{(M)}(\widetilde{\vx})\Big)\in\R^{P_M},
    \qquad P_M:=\sum_{k=0}^M (d+1)^{k}.
\end{equation}
\end{definition}
Throughout, the iterated integrals are interpreted in the Stratonovich sense unless otherwise specified.

Because the full signature coordinates in \cref{eq:sm-full-signature} can exhibit collinearity induced by their algebraic structure, the signature pruning technique can be applied to remove feature redundancy. Formally, it embeds each path through a fixed feature map $\vpsi:\mathcal X\to\R^P$ shared across all tasks, constructed from the pruned signature coordinates up to level $M$:
\begin{equation}\label{eq:sm-feature-map}
\begin{aligned}
    \vpsi(\vx):=\overline{\Sig}_M(\widetilde{\vx})
    :=\Big(1,\Big(
       \bigl[\Sig^{(k)}(\widetilde{\vx})\bigr]_{i_1,\ldots,i_k}
       \Big)_{\substack{1\leq k\leq M,\ i_k\neq0}}\Big)\in\R^P,\
    P:=(d+1)^M.
\end{aligned}
\end{equation}
Crucially, there is no loss of expressiveness from this pruning, as every coordinate of the full truncated signature $\Sig_M(\widetilde{\vx})$ lies in the linear span of $\overline{\Sig}_M(\widetilde{\vx})$ \citep{guo2026signature}.

In practice, 
for discretely observed data, we linearly interpolate between consecutive observations before time augmentation.  The augmented path is then piecewise linear, and Chen's identity~\citep{chen1954iterated} expresses its signature coordinates as polynomials in the observations.  
\section{Theory of Signature-Based In-Context Learning}
\label{sec:sm-main}

\subsection{In-Context Learning in Path Space: Problem Formulation}

In ICL,  the target function is unknown and varies across tasks, so the transformer must learn from the demonstrations in context to predict the task-specific label for the query, while keeping its parameters fixed. 

In the standard formulation of ICL, each demonstration pairs a finite dimensional input feature vector with its label:
\begin{equation}
(\vx_1,Y_1),\ldots,(\vx_N,Y_N),\vx_{N+1}, \qquad \vx_i\in\R^d.
\end{equation}
The goal is to predict the label corresponding to the  query $\vx_{N+1}$.

This formulation can be generalized to path inputs.
Let $\mathcal W$ denote the task space and $\mathcal F:=\{f_w:w\in\mathcal W\}$ the corresponding class of target functionals $f_w:\mathcal X\to\R$ on a standard Borel path space $\mathcal X$. 
Each task $w\in\mathcal W$ selects $f_w\in\mathcal F$ and may also govern the distribution of the input process.
Within each prompt, demonstrations and the query share the
same unknown task $w$. 
Let $\vZ=(\vZ_t)_{t\geq0}$ be an input process in $\R^d$. For $i\geq1$, we define the $i$-th demonstration path by restricting $\vZ$ to $[i-1,i]$ and reparametrizing time onto $[0,1]$:
\begin{equation}\label{eq:sm-segments}
    \vX_i = (\vZ_{i-1+t})_{0\leq t\leq1},\qquad i=1,2,\ldots.
\end{equation}
Each segment $\vX_i$ takes values in $\mathcal X$. For an integer $N\geq1$, the labels are
\begin{equation}\label{eq:sm-labels}
    Y_i = f_w(\vX_i) + \xi_i, \qquad i=1,\ldots,N+1,
\end{equation}
where $\xi_i$ denotes label noise. The full prompt presented to the
transformer is $(\vX_i,Y_i)_{i=1}^N$ together with the query $\vX_{N+1}$.
For each task $w$, write $\P_w$ for the joint law of the input process and label noises, and $\E_w$ for the corresponding expectation.

In \cref{sec:financial-payoff-icl}, we will present and analyze an example, where each demonstration
pairs a price path with the payoff of an undisclosed path-dependent option. All demonstrations in a prompt use the same option. Given a new price path as the query, the model predicts its payoff from that option.

\paragraph{Prompt format.}
Given $(\vX_i,Y_i)_{i=1}^N$ and the query $\vX_{N+1}$, the input sequence, with token dimension $D=2P+3$, is
\begin{equation}\label{eq:sm-prompt-encoding}
 \mH^{(0)}
 =
 \left(
 \begin{array}{c|ccc|c}
 \vzero_P & \vpsi(\vX_1) & \cdots & \vpsi(\vX_N) & \vpsi(\vX_{N+1})\\
 0   & Y_1       & \cdots & Y_N       & 0\\
 \vzero_P & \vzero_P       & \cdots & \vzero_P       & \vzero_P\\
 0   & 1         & \cdots & 1         & 0\\
 0   & 0         & \cdots & 0         & 1
 \end{array}
 \right)^{\!\top}
 \in\R^{(N+2)\times(2P+3)}.
\end{equation}
Here $\vh_0^{(0)}$ is the fixed zero anchor, each demonstration $(\vX_i,Y_i)$ corresponds to token $\vh_i^{(0)}$ for $1\leq i\leq N$, and the query $\vX_{N+1}$ corresponds to $\vh_{N+1}^{(0)}$. Each example or query token holds its signature features and label (zero for the query).
The next $P$ coordinates are initialized to zero and reserved for the coefficient iterate. The final two coordinates indicate labeled examples and query tokens, respectively. 
\paragraph{Read out operator.}
Given an input sequence $\mH^{(0)}$, the transformer yields the prediction
$\widehat y_{\rm TF}=\mathrm{read}_y(\TF_\vtheta(\mH^{(0)}))$, where $\mathrm{read}_y$ is a linear readout operator that maps the final output matrix to a scalar by extracting the $(P+1)$-th coordinate of the final token:
\begin{equation}\label{eq:sm-readout-def}
    \widehat y_{\rm TF}=\mathrm{read}_y(\TF_\vtheta(\mH^{(0)}))
    =[(\TF_\vtheta(\mH^{(0)}))_{N+1}]_{P+1}.
\end{equation}
The population squared loss is thus
\begin{equation}\label{eq:icl-loss}
    \mathcal L_w(\vtheta)
    =\E_w\big[(\widehat y_{\rm TF}-Y_{N+1})^2\big],
\end{equation}
where the expectation is over prompts and query responses under the fixed task $w$.

The training objective of $\vtheta$ is to reduce this loss across tasks.
At inference, the parameters $\vtheta$ remain fixed, and the model uses the demonstrations to predict the task-specific label for the query.

\subsection{Universal Nonlinearity of Signature Transform for ICL}
\label{sec:sm-results}

We now present a uniform signature approximation over a class of continuous functionals.
This is the universal nonlinearity property of the signature transform for the ICL setting.

\begin{lemma}[Uniform signature approximation]
\label{lem:sig-ua}
Let $\mathcal K\subset\mathcal X$ be a set of paths starting at zero,
compact in the $p$-variation metric for some $p\geq1$, and let
$\mathcal F$ be a class of functionals whose restrictions to
$\mathcal K$ are continuous and totally bounded in the supremum norm
$\norm{f}_\infty:=\sup_{\vx\in\mathcal K}|f(\vx)|$.
For any $\epsilon>0$, there exists a truncation level $M$ and coefficient vectors
$\vbeta_f\in\R^{P_M}$, $f\in\mathcal F$, such that
\begin{equation}\label{eq:uniform-sig-approx}
    \sup_{f\in\mathcal F}\sup_{\vx\in\mathcal K}
    \abs{f(\vx)-\vbeta_f^\top\Sig_M(\widetilde{\vx})}\leq\epsilon.
\end{equation}
\end{lemma}
The truncation level $M$
is uniform over $\mathcal F$, while the coefficients may depend on $f$.
The lemma indicates that path-dependent target functions can be approximated by linear combinations of a common set of signature features.
Consequently, the original ICL problem is reduced to fitting a task-specific linear model within a transformer.
The proof of Lemma~\ref{lem:sig-ua} is given in \cref{app:proof_sig_approx}. 

\paragraph{Signature level and truncation error.}

Generally speaking, the nonlinearity property of the signature is a Stone--Weierstrass-type result that guarantees uniform approximation but does not specify an explicit rate relating $\epsilon$ to $M$.
Recently, \cite{horvath2026fast} established, under additional smoothness assumptions,  an approximation error for Itô diffusions of order $\epsilon = \mathcal{O}(M^{-2\nu})$, where $\nu > 0$ measures the smoothness of the functional.
\citet{schenck2026finitesample} gives finite-sample mean squared error
bounds for truncated expected signatures from dependent observations,
including fractional Ornstein--Uhlenbeck processes.

Empirically, low truncation levels are shown to  provide good approximations to path functionals in applications.
In finance, terminal and Asian payoffs admit exact linear representations in augmented signatures truncated up to  levels three \citep{perezarribas2018derivatives}. \citet{guo2025consistency} fits European, Asian, and lookback option payoffs using signatures truncated at level six, and \citet{perezarribas2018derivatives} approximates derivative prices under the Black--Scholes model using level four.
\citet{lyons2019numerical} also obtains accurate exotic derivative prices at
level five, and \citet{futter2023signature} approximates an MACD momentum
strategy by a linear signature functional at level three, achieving
$R^2=89.2\%$.
Beyond derivatives, \citet{gu2024transportation} combine signature
features with Lasso regression to forecast freight transportation rates, with signatures
truncated at level three.

\subsection{Universal Expressivity of Signature-ICL and Error-bound Analysis}

\paragraph{In-context Lasso.}
By the universal signature approximation (\cref{lem:sig-ua}), the original ICL problem is reduced to fitting a task-specific linear model within a transformer. 

To exploit the sparsity in the high-dimensional signature
space~\citep{guo2025consistency}, we use the Lasso estimator.
Let $\mPsi\in\R^{N\times P}$ have rows of the pruned signature $\vpsi(\vX_i)^\top$ (\cref{eq:sm-feature-map}), 
$\vy=(Y_1,\ldots,Y_N)^\top$, and define the Lasso objective
\begin{equation}\label{eq:sm-lasso-objective}
    J_\lambda(\vu)
    :=\frac1{2N}\norm{\mPsi\vu-\vy}_2^2+\lambda\norm{\vu}_1,
    \qquad \lambda>0.
\end{equation}
Now, define the corresponding empirical
Gram matrix 
\begin{equation}
    \mG_N:=N^{-1}\mPsi^\top\mPsi=\frac{1}{N}\sum_{i=1}^N\vpsi(\vX_i)\vpsi(\vX_i)^\top.
\end{equation}
We will show that transformers (\cref{def:transformer}) with signature embeddings have universal expressivity to solve ICL problems in path space.

The analysis relies on several assumptions, which will be further discussed in Section \ref{sec:sm-examples}.
All constants in the following are common across tasks.
\begin{Assumption}[Bounded moments]
\label{ass:sm-bounded}
For any task $w$ and $i\geq1$,
$\E_w\norm{\vpsi(\vX_i)}_2^4\leq B^4$.
\end{Assumption}

\begin{Assumption}[Sparse signature approximation]
\label{ass:sm-sparse}
For some $\epsilon\geq0$ and integer $s<P$,
any target $f_w\in\mathcal F$, $w\in\mathcal W$, admits a vector $\vbeta_w^\star\in\R^P$
satisfying, for every $\vX\in \mathcal{X}$,
\begin{equation}\label{eq:sm-sparse-approximation}
    \abs{f_w(\vX)-\vpsi(\vX)^\top\vbeta_w^\star}\leq\epsilon,
    \quad \|\vbeta_w^\star\|_0\leq s.
\end{equation}
The coefficients satisfy $\norm{\vbeta_w^\star}_2\leq R$.
\end{Assumption}
Because the 
dimension of the signature grows exponentially with the
truncation depth, 
 sparsity on the coefficient vector are commonly imposed for
statistical analyses of signature regression, for instance,  in \citep{horvath2026fast} for signature approximation of the Ornstein--Uhlenbeck process
 and in \citep{guo2026tradable} for sparse It\^o signature regression of
derivative hedging. 

For a given task $w$, let $S:=\supp(\vbeta_w^\star)$ denote the support of $\vbeta_w^\star$.

\begin{Assumption}[Restricted strong convexity (RSC)]
\label{ass:sm-rsc}
Let
\begin{equation}\label{eq:sm-population-covariance}
    \E_w[\mG_N]
    =\frac1N\sum_{i=1}^N\E_w[\vpsi(\vX_i)\vpsi(\vX_i)^\top],\quad
    \mathcal C(S):=
    \{\vv\in\R^P:\norm{\vv_{S^c}}_1\leq4\norm{\vv_S}_1\}.
\end{equation}
There is a common $\alpha>0$ such that, for every task $w$,
\begin{equation}\label{eq:sm-population-rsc}
    \vv^\top\E_w[\mG_N] \vv\geq2\alpha\norm \vv_2^2,
    \qquad \forall\ \vv\in\mathcal C(S).
\end{equation}
\end{Assumption}

RSC is a standard condition for Lasso analysis \citep{negahban2012unified}. Note that it requires curvature of $\E_w[\mG_N]$ only on the cone $\mathcal C(S)$ instead of the whole space $\R^P$.
For smooth uniformly elliptic diffusions, linear independence of the pruned time-augmented signature coordinates implies positive definiteness of their Gram matrix at each fixed truncation level \citep{bourdon2026linear}.

\begin{Assumption}[Concentration]
\label{ass:sm-concentration}
There exists an absolute constant $c>0$ such that, for any task $w$, the empirical Gram matrix satisfies
\begin{equation}\label{eq:sm-prompt-event}
    \P_w\big(
       \sup_{\vv\in\mathcal C(S),\norm{\vv}_2\leq1}
       |\vv^\top(\mG_N-\E_w[\mG_N])\vv|\leq\alpha
    \big)\geq1-c(PN)^{-4}.
\end{equation}
In addition, the label noise has conditional mean zero and is $\sigma$-sub-Gaussian:
\begin{equation}\label{eq:sm-noise-mgf}
    \E_w[e^{t\xi_i}\mid \vX_1,\ldots,\vX_i,\xi_1,\ldots,\xi_{i-1}]
    \leq e^{\sigma^2t^2/2},
    \qquad t\in\R,\quad i=1,\ldots,N.
\end{equation}
\end{Assumption}

The concentration assumption ensures that the empirical Gram matrix
$\mG_N$ inherits the RSC property of $\E_w[\mG_N]$. 
It can be verified
for several standard stochastic processes, as shown in \cref{sec:sm-examples}.

Now we are ready to present our main results, a universal ICL approximation theory of signature-based transformers.
\begin{thm}[Universal ICL of path functionals]
\label{thm:sm-main}
Suppose \cref{ass:sm-bounded,ass:sm-sparse,ass:sm-rsc,ass:sm-concentration}
hold.
For any $\epsilon_{\rm TF}>0$, there exists a transformer $\TF_\vtheta(\cdot)$ and a constant $C$ such that, for any $w\in\mathcal W$,
\begin{equation}\label{eq:sm-general-rate}
    \E_w\bigl(\widehat y_{\rm TF}-f_w(\vX_{N+1})\bigr)^2
    \leq C\Big(
       \epsilon^2+\frac{\sigma^2s\log(PN)}{N}
       +\frac{1}{PN^2}
    \Big)+\epsilon_{\rm TF}^2.
\end{equation}
Here $\widehat y_{\rm TF}=\mathrm{read}_y(\TF_\vtheta(\mH^{(0)}))$,
with $\mH^{(0)}$ defined in \cref{eq:sm-prompt-encoding}.
The transformer implements
in-context signature Lasso: it outputs coefficients
$\norm{\widehat{\vbeta}}_\infty\leq R$ satisfying
$\E_w\bigl(\widehat y_{\rm TF}
 -\langle\overline{\Sig}_M(\widetilde{\vX}_{N+1}),\widehat{\vbeta}\rangle\bigr)^2\leq\epsilon_{\rm TF}^2$ and
$\E_w[J_\lambda(\widehat{\vbeta})-\min_{\norm{\vu}_\infty\leq R}J_\lambda(\vu)]\leq\epsilon_{\rm TF}$ for some penalty $\lambda$.

Moreover, the transformer has $\mathcal{O}(\epsilon_{\rm TF}^{-3})$ layers, with each layer comprising two attention heads of dimension $P+1$ and an
MLP of width $4P+1$, with $P=(d+1)^M$.
\end{thm}

\Cref{thm:sm-main} gives an explicit construction of a causal softmax transformer that uses the signature as path embeddings and simulates in-context Lasso:  two softmax heads per layer and an MLP that performs soft-thresholding and projection. 

It establishes an in-context prediction guarantee for general nonlinear, path-dependent target functions, extending the scope of existing in-context linear-regression constructions \citep{bai2023transformers,hu2025softmax}. A single signature-based transformer learns any $f_w\in\mathcal F$ from the demonstrations in context without retraining its parameters. 
The bound in \cref{eq:sm-general-rate} consists of   several sources of error: $\epsilon^2$ captures the signature approximation error,
$\sigma^2s\log(PN)/N$ is the canonical Lasso estimation rate, and
$1/(PN^2)$ is a tail error term from outside the truncation event. Finally, $\epsilon_{\rm TF}^2$ represents the approximation error arising from the transformer construction.

For exact signature representations ($\epsilon=0$), sufficiently small $\epsilon_{\rm TF}$, and label noise $\sigma^2\geq \mathcal{O}(1/PN)$, the theorem yields
\begin{equation}\label{eq:sm-canonical-rate}
    \E_w\bigl(\widehat y_{\rm TF}-f_w(\vX_{N+1})\bigr)^2
    \leq C\frac{\sigma^2s\log(PN)}{N}.
\end{equation}
This matches the classical Lasso rate up to logarithmic factors \citep{bickel2009simultaneous,negahban2012unified}.

\subsection{Assumptions on Bounded Moments and Concentration Inequalities}
\label{sec:sm-examples}

We now verify \cref{ass:sm-bounded} and the empirical RSC bound
\cref{eq:sm-prompt-event} for two classes of processes commonly adopted in financial applications: geometrically mixing diffusions and fractional Ornstein--Uhlenbeck processes. 
Beyond their role in our ICL analysis, these concentration guarantees are of independent interest for high-dimensional signature regression on dependent paths.
The proofs are given in \cref{app:verify_moments}.

\subsubsection{Diffusion Processes with Geometric Mixing}
\label{sec:sm-diffusions}

We first consider dissipative diffusion processes, which are widely studied in short-rate modeling \citep{vasicek1977equilibrium},
commodity pricing \citep{schwartz1997stochastic},
portfolio optimization
\citep{guasoni2012portfolios,liang2017representation},
and dynamic risk measurement \citep{chong2019ergodic}.
Let $\vZ$ be a diffusion process on $\R^d$ satisfying
\begin{equation}\label{eq:sm-brownian-sde}
    \dif \vZ_t=\vb(\vZ_t)\,\dif t+\msigma(\vZ_t)\,\dif \vW_t,
    \qquad \vZ_0=\vz_0,
\end{equation}
where $\vW$ is a standard $d$-dimensional Brownian motion.
\begin{prop}
\label{prop:sm-brownian-sde}
Suppose that $\vb$ and $\msigma$ are globally Lipschitz and satisfy, for all $\vz\in\R^d$,
\begin{equation}\label{eq:sm-sde-coefficients}
\langle\vz,\vb(\vz)\rangle
\leq -m\norm{\vz}_2^2+\gamma,\qquad
\underline c\,\mI_d\preceq\msigma(\vz)\msigma(\vz)^\top
    \preceq\bar c\,\mI_d,
\end{equation}
where $m,\underline c,\bar c,\gamma>0$ are constants.
Then the signature features have uniformly bounded fourth moments.
Moreover, with probability at least $1-\delta$,
\begin{equation}\label{eq:sm-diffusion-rates}
\begin{split}
    \sup_{\vv\in\mathcal C(S),\norm{\vv}_2\leq1}
       |\vv^\top(\mG_N-\E_w[\mG_N])\vv|
    \leq C\Big(
       \sqrt{\frac{s\log(P/\delta)\log(N/\delta)}{N}}
       +\frac{s\log(P/\delta)[\log(N/\delta)]^{M+1}}{N}
    \Big).
\end{split}
\end{equation}
\end{prop}
Dissipativity with uniform ellipticity leads to geometric ergodicity and thus geometric mixing of successive unit-time paths
\citep{mattingly2002ergodicity}.
Together with the signature tail estimates and Bernstein's inequality, we have the restricted empirical RSC in \cref{eq:sm-diffusion-rates}.
In particular, \cref{ass:sm-bounded,eq:sm-prompt-event} hold if
\begin{equation}\label{eq:sm-diffusion-sample-size}
    N\geq C s[\log P]^{M+2}.
\end{equation}

\paragraph{CIR short-rate process.}
The classical Cox--Ingersoll--Ross (CIR) short-rate model
\citep{cox1985theory} satisfies
\begin{equation}\label{eq:sm-cir-process}
    \dif Z_t=\kappa(\mu-Z_t)\,\dif t+\nu\sqrt{Z_t}\,\dif W_t,
    \qquad Z_0=z_0,
\end{equation}
where $\kappa,\mu,\nu>0$, and
$z_0\geq0$.
Its diffusion coefficient does not satisfy uniform ellipticity,
so \cref{prop:sm-brownian-sde} does not apply. Nevertheless, its transition kernel is geometrically ergodic \citep{jin2016exponential}, and we can verify the conditions directly.

\begin{prop}
\label{prop:sm-cir-concentration}
For the CIR process, the signature features have uniformly bounded
fourth moments. Moreover, with probability at least
$1-\delta$,
\begin{equation}\label{eq:sm-cir-rates}
\begin{split}
    \sup_{\vv\in\mathcal C(S),\norm{\vv}_2\leq1}
       |\vv^\top(\mG_N-\E_w[\mG_N])\vv|
    \leq C\Big(
       \sqrt{\frac{s\log(P/\delta)\log(N/\delta)}{N}}
       +\frac{s\log(P/\delta)[\log(N/\delta)]^{2M+1}}{N}
    \Big).
\end{split}
\end{equation}
\end{prop}
Therefore \cref{ass:sm-bounded,eq:sm-prompt-event} holds for the CIR process if $N\geq Cs[\log P]^{2M+2}$.

\subsubsection{Paths with Long-Range Dependence}
\label{sec:sm-fractional-gaussian}

Fractional Gaussian factors appear in long-memory stochastic volatility
models \citep{comte1998long} and rough log-volatility models
\citep{gatheral2018volatility}. We focus on the long-range regime
$H>1/2$ but treat all Hurst parameters for comparison.

Consider the one-dimensional fractional Ornstein--Uhlenbeck (OU) process
\begin{equation}\label{eq:sm-fou-process}
    \dif Z_t=-\kappa Z_t\,\dif t+\nu\,\dif W_t^H,
    \qquad Z_0=z_0,
\end{equation}
where $W^H$ is a standard fractional Brownian motion with
$H\in(0,1)$ and $\kappa,\nu>0$.
For $H>1/2$, the covariance of its stationary counterpart decays at rate
$t^{2H-2}$ and is not integrable \citep{cheridito2003fractional}.
Gaussian covariance estimates give the following bounds for all
$H\in(0,1)$.
\begin{prop}
\label{prop:sm-fou-concentration}
For the fractional OU process, the signature features have uniformly
bounded fourth moments.
Moreover, with probability at least $1-\delta$,
\begin{equation}\label{eq:sm-fou-rates}
    \sup_{\vv\in\mathcal C(S),\norm{\vv}_2\leq1}
       |\vv^\top(\mG_N-\E_w[\mG_N])\vv|
    \leq C[\log(N/\delta)]^{M-1/2}
       \sqrt{\frac{s\log(P/\delta)}{N^{2-\max(2H,1)}}}.
\end{equation}
\end{prop}
The constants depend only on $M,H,\kappa,\nu$, and $z_0$, so
\cref{ass:sm-bounded,eq:sm-prompt-event} hold if
\begin{equation}\label{eq:sm-fou-sample-size}
    N\geq C\bigl(s[\log P]^{2M}\bigr)^{1/(2-\max(2H,1))},
\end{equation}
where $C$ also depends on $\alpha$.
For $H>1/2$, concentration is governed by $N^{2-2H}$, so long-range
dependence increases the required sample size; for $H\leq1/2$ the rate
is $N$, and $H=1/2$ gives the ordinary OU process.

\section{Proof of  \Cref{thm:sm-main} }
\label{sec:sm-proof-overview}

Assuming the sparse signature approximation, which  reduces the non-linear, path-dependent prediction problem to sparse linear regression in the signature feature space,   we can decompose the prediction error as
\begin{equation}\label{eq:sm-error-decompose}
    \begin{aligned}
        \widehat y_{\rm TF}-f_w(\vX_{N+1})
        &= \underbrace{\bigl(\widehat y_{\rm TF}-\vpsi(\vX_{N+1})^\top\widehat\vbeta\bigr)}_{\text{I. Transformer construction}}  + \underbrace{\vpsi(\vX_{N+1})^\top(\widehat\vbeta-\vbeta_w^\star)}_{\text{II. Lasso}} \\
        &\quad
        + \underbrace{\bigl(\vpsi(\vX_{N+1})^\top\vbeta_w^\star-f_w(\vX_{N+1})\bigr)}_{\text{III. Signature approximation}}.
    \end{aligned}
\end{equation}
Here $\vbeta_w^\star$ is the ground-truth coefficient in signature space and $\widehat\vbeta$ is an estimator of signature in-context Lasso problem.

In this section, we first outline how the constructed transformer performs in-context Lasso (I), then establish the statistical rate of the signature Lasso estimator (II), 
and finally combine these ingredients to prove our main theorem.

\subsection{Step I: Construction of the Transformer}

Recall that the  transformer takes the prompt format \cref{eq:sm-prompt-encoding}
as input and the $\ell$-th layer stores the current coefficient estimate $\vv^{(\ell)}$ in the last token, i.e., 
\begin{equation}
    \vv^{(\ell)}=[\mH_{N+1}^{(\ell)}]_{P+2:2P+1}\in\R^P.
\end{equation}
It approximately minimizes the Lasso objective over $[-R,R]^P$, which contains $\vbeta_w^\star$ for every task 
\cref{ass:sm-sparse}. The following lemma summarizes this designated mechanism.
\begin{lemma}[Transformer solves in-context Lasso]
\label{lem:sm-transformer-lasso-main}
Fix a truncation radius $K$. For any $\gamma>0$, there is a transformer with $L+1$ layers and a step size $\eta>0$ satisfying the following for any task $w$ when $\max_i\norm{\vpsi(\vX_i)}_2\leq K$ and $\max_i|Y_i|\leq K$. 
\begin{enumerate}[label=(\arabic*),leftmargin=*]
\item It contains $L$ optimization layers, and each approximates one proximal-gradient step for $J_\lambda(\cdot)$ in \cref{eq:sm-lasso-objective}:
\begin{equation}
    \Big\|\vv^{(\ell+1)}-
       \operatorname{proj}_{[-R,R]^P}\soft_{\eta\lambda}\Big(
          \vv^{(\ell)}-\frac{\eta}{N}\mPsi^\top
             (\mPsi\vv^{(\ell)}-\vy)
       \Big)\Big\|_2\leq\gamma,
    \qquad 0\leq\ell<L.
\end{equation}
Here $\soft_t(u)=\operatorname{sign}(u)(|u|-t)_+$.
Specifically, the attention block approximates the squared-loss gradient step, while the MLP
applies nonlinear soft-thresholding and projection. 
\item It contains one label readout layer, producing coefficients $\vv=\vv^{(L)}$ and prediction $\widehat y_{\rm TF}$ such that
\begin{equation}\label{eq:sm-transformer-output-error}
    J_\lambda(\vv)\leq\min_{\vu\in[-R,R]^P}J_\lambda(\vu)+\gamma,
    \qquad
    |\widehat y_{\rm TF}-\vpsi(\vX_{N+1})^\top\vv|\leq\gamma.
\end{equation}
\item Each layer consists of two attention heads with head dimension $P+1$ and an MLP block of width $4P+1$, and $L=\mathcal{O}(K^2R^2P/\gamma)$.
\end{enumerate}
Moreover, for any input prompt (without requiring the bounds on $\norm{\vpsi(\vX_i)}_2$ and $\norm{Y_i}$), it holds that $\norm{\vv}_\infty\leq R$, $|\widehat y_{\rm TF}|\leq|\vpsi(\vX_{N+1})^\top\vv|\leq R\sqrt P\norm{\vpsi(\vX_{N+1})}_2$.
\end{lemma}

\begin{proof}[Proof Sketch (for details see Appendix~\ref{sec:transfer-construction})]
Recall the Lasso objective
$J_\lambda(\vu):=\frac1{2N}\sum_{i=1}^N(\vpsi(\vX_i)^\top\vu-Y_i)^2+\lambda\norm{\vu}_1$.
We write a token embedding as $\vh=[\vx;y;\vv;b_{\rm d};b_{\rm q}]$, where
$\vx\in\R^P$ is its feature vector and $b_{\rm d}$ and $b_{\rm q}$
indicate demonstration and query tokens, respectively.
We encode the path-label pairs and the query, together with a zero anchor, as
\begin{equation}
\begin{aligned}
    \vh_0&=[\vzero_P;0;\vzero_P;0;0],\\
    \vh_i&=[\vpsi(\vX_i);Y_i;\vzero_P;1;0],\qquad 1\leq i\leq N,\\
    \vh_{N+1}&=[\vpsi(\vX_{N+1});0;\vv;0;1].
\end{aligned}
\end{equation}
The coefficient vector $\vv$ is initially zero and stores the
successive coefficient estimates. 
We start with input $\vv^{(0)}=\vzero_P$ in the query token, apply $L$ optimization layers, and append one label readout layer.

\paragraph{Optimization layers.}
The central mechanism is that the attention operator naturally emulates the empirical squared-loss gradient through its bilinear inner-product structure, as established in \cref{lem:sm-softmax-gradient}. 

Here is the key idea. For current coefficients $\vv$, let
$r_i=\vpsi(\vX_i)^\top\vv-Y_i$ denote the residual. The desired gradient update is
\begin{equation}
    \vv\longmapsto\vv-\frac{\eta}{N}\sum_{i=1}^N
       r_i\vpsi(\vX_i).
\end{equation}
In an attention block, the query state $\vh_{N+1}$ updates itself by taking a weighted sum of value vectors extracted from the context states $\vh_i$, where the weights are determined by the inner products $\langle\mQ^\top\vh_{N+1},\mK^\top\vh_i\rangle$. Thus, by placing the current coefficient vector $\vv$ in $\vh_{N+1}$ as query and the context $(\vpsi(\vX_i), Y_i)$ in $\vh_i$ as key, the inner product evaluates the residual $r_i$ directly. Specifically, for parameters $A,\tau>0$, the two-head projections are chosen as
\begin{equation}
\begin{aligned}
    &\mQ_\pm^\top\vh_{N+1}=[\vv;1]\in\R^{P+1},\qquad
    \mK_\pm^\top\vh_i=[\pm\tau\vpsi(\vX_i);\mp\tau Y_i-A]\in\R^{P+1},\\
    &\mV_\pm^\top\vh_i=\mp\frac{\eta e^A}{2\tau N}
       [\vzero_P;0;\vpsi(\vX_i);0;0].
\end{aligned}
\end{equation}
Appending the constant $1$ to $\mQ_\pm^\top\vh_{N+1}$ allows the inner product with $\mK_\pm^\top\vh_i$ to compute the affine residual $\langle\mQ_\pm^\top\vh_{N+1},\mK_\pm^\top\vh_i\rangle = -A\pm\tau r_i$, while the linear map $\mV_\pm^\top$ places the feature vector $\vpsi(\vX_i)$ into the coordinates corresponding to $\vv$ with opposite signs across the two heads.

To see how the two heads combine, note that the zero anchor $\vh_0$ contributes $1$ to each softmax denominator. Taking $A$ large drives both denominators toward $1$, while the factor $e^A$ in the value vectors compensates for $e^{-A}$ in the demonstration weights. The combined coefficient vector update to $\vv$ is then close to
\begin{equation}
    \vv-\frac{\eta}{N}\sum_{i=1}^N
       \frac{e^{\tau r_i}-e^{-\tau r_i}}{2\tau}\vpsi(\vX_i)
    \longrightarrow
    \vv-\frac{\eta}{N}\sum_{i=1}^N r_i\vpsi(\vX_i)
    \qquad\text{as }\tau\downarrow0.
\end{equation}


The subsequent MLP block applies soft-thresholding at level $\eta\lambda$ and clips the coefficients to $[-R,R]^P$, completing the proximal-gradient step as in  \cref{lem:sm-threshold}. Note that these coordinatewise operations are piecewise linear and can be represented exactly by the MLP with ReLU activations and width $4P+1$.
Both blocks preserve the demonstration tokens, so every optimization layer uses the same  observations.

\paragraph{Label readout layer.}
Now the final layer keeps $\vv^{(L)}$ in the coefficient coordinates. Its two attention heads approximate the inner product $\vpsi(\vX_{N+1})^\top\vv^{(L)}$ and outputs $\widehat y_{\rm TF}$. Setting the MLP's weights to zero makes its residual map the identity. The transformer thus has $L$ optimization layers and one readout layer to yield $\widehat{y}_{\rm TF}$, as detailed in \cref{lem:sm-readout}.

\paragraph{Optimization bound.}
Finally, these components are stacked across layers to solve in-context Lasso up to any prescribed optimization tolerance. Specifically, with step size $\eta=(1+K^2)^{-1}$, the standard proximal-gradient convergence rate \citep{beck2017first} gives an objective gap of at most $\gamma/2$ after $L=\mathcal{O}(K^2R^2P/\gamma)$ iterations. Because the exact proximal update is nonexpansive, the iterate error after $L$ layers is bounded by $L$ times the per-layer approximation error.
Choosing the per-layer attention error sufficiently small ensures that the accumulated discrepancy contributes at most $\gamma/2$ to the objective, hence the \cref{eq:sm-transformer-output-error}, with detailed proofs given in \cref{app:proof_transformer}.
\end{proof}

\subsection{Step II: Statistical Rate of Lasso on Signature Features}

The following lemma shows that the approximate in-context Lasso solution
produced by our constructed transformer accurately recovers the underlying
sparse coefficients, even under dependent path observations.
\begin{lemma}[Lasso oracle inequality for dependent data]
\label{lem:sm-prompt-event-main}
For any task $w\in\mathcal W$, suppose
\cref{ass:sm-bounded,ass:sm-sparse,ass:sm-rsc,ass:sm-concentration}
hold. With probability at least $1-C(PN)^{-4}$ under $\P_w$, the following
statement holds. Let $\lambda>0$ satisfy
$\lambda\geq C\sqrt{B^2+\alpha}\big(\epsilon+
\sigma\sqrt{\log(PN)/N}\big)$ and let $\gamma\geq0$.
If $\vv\in\R^P$ satisfies
$J_\lambda(\vv)\leq J_\lambda(\vbeta_w^\star)+\gamma$, then
\begin{equation}\label{eq:sm-prompt-oracle}
    \norm{\vv-\vbeta_w^\star}_2
    \leq C\Big(\frac{\sqrt{s}\lambda}{\alpha}
         +\frac{\gamma}{\lambda}\Big).
\end{equation}
\end{lemma}

The lemma converts the objective gap guarantee of \cref{lem:sm-transformer-lasso-main} into an $\ell_2$ error bound relative to the target sparse signature coefficients $\vbeta_w^\star$.

We emphasize that this bound applies to any $\gamma$-suboptimal
solution of the Lasso objective, and the numerical optimization tolerance adds only a penalty of order $\gamma/\lambda$.

In contrast to  prior analyses of exact Lasso estimators for time series use $L^q$ functional dependence measures to quantify decaying temporal correlations \citep{han2020dependent,adamek2023lasso}, our result here takes a different approach by directly assuming that the empirical Gram matrix of the concentrates around its expectation on the sparsity cone, as stated in \cref{eq:sm-prompt-event}.

\subsection{The Error-bound Analysis}

We now finish the proof of  \Cref{thm:sm-main}, by  combining the transformer construction in \cref{lem:sm-transformer-lasso-main} with the Lasso bound in \cref{lem:sm-prompt-event-main} and the error decomposition in \cref{eq:sm-error-decompose}.

 Let us first set the regularization parameter $\lambda$ and optimization tolerance
$\gamma$ as
\begin{equation}\label{eq:sm-main-transformer-tuning}
    \lambda:=C\sqrt{B^2+\alpha}\Big(
       \epsilon+\sigma\sqrt{\frac{\log(PN)}{N}}\Big),
    \qquad
    \gamma:=\frac{c_1\epsilon_{\rm TF}\lambda}{B+\lambda},
\end{equation}
for a sufficiently large constant $C>0$ and sufficiently small constant
$c_1>0$.
Fix a task $w$, then \cref{lem:sm-prompt-event-main} yields an event $\mathcal E$ with
$\P_w(\mathcal E^c)\leq C(PN)^{-4}$ on which the oracle bound
\cref{eq:sm-prompt-oracle} holds.

\paragraph{Truncation and optimization bounds.}
Since $|f_w(\vX_i)|\leq R\norm{\vpsi(\vX_i)}_2+\epsilon$,
\cref{ass:sm-bounded} and the sub-Gaussian noise condition imply
$\sup_{i\leq N}\E_w|Y_i|^4\leq C(R^4B^4+\epsilon^4+\sigma^4)$. For a truncation radius $K>0$, define the truncation event
\begin{equation*}
    \mathcal E_K:=
    \left\{\max_{i\leq N+1}\norm{\vpsi(\vX_i)}_2\leq K,\quad
           \max_{i\leq N}|Y_i|\leq K\right\},
\end{equation*}
which satisfies $\P_w(\mathcal E_K^c)=\mathcal{O}(K^{-4})$ by Markov's inequality.
Let $\TF_\vtheta$ be the transformer from
\cref{lem:sm-transformer-lasso-main} with truncation radius $K$ and
optimization tolerance $\gamma$, and write $\widehat{\vbeta}:=\vv^{(L)}$ for its
extracted coefficient vector. 

 For every prompt, the bounds
$\norm{\widehat\vbeta}_\infty\leq R$ and
$|\widehat y_{\rm TF}|\leq R\sqrt P\norm{\vpsi(\vX_{N+1})}_2$
from \cref{lem:sm-transformer-lasso-main} imply
\begin{equation}\label{eq:sm-global-prediction-bound}
\begin{aligned}
    \bigl(\widehat y_{\rm TF}-f_w(\vX_{N+1})\bigr)^2
    &\leq2R^2(\sqrt P+1)^2\norm{\vpsi(\vX_{N+1})}_2^2
       +2\epsilon^2=:U_{\rm pred},\\
    J_\lambda(\widehat\vbeta)-\min_{\norm{\vu}_\infty\leq R}J_\lambda(\vu)
    &\leq\frac1N\sum_{i=1}^N
       \bigl(R^2P\norm{\vpsi(\vX_i)}_2^2+Y_i^2\bigr)+\lambda RP
    =:U_{\rm obj},\\
    \bigl(\widehat y_{\rm TF}-\vpsi(\vX_{N+1})^\top\widehat\vbeta\bigr)^2
    &\leq4R^2P\norm{\vpsi(\vX_{N+1})}_2^2
    =:U_{\rm read}.
\end{aligned}
\end{equation}
By \cref{ass:sm-bounded}, each
$U\in\{U_{\rm pred},U_{\rm obj},U_{\rm read}\}$ has a uniformly bounded
second moment. Therefore
$\E_w\bigl[U\mathbf1_{\mathcal E_K^c}\bigr]
\leq(\E_w U^2)^{1/2}\P_w(\mathcal E_K^c)^{1/2}=\mathcal{O}(K^{-2})$ uniformly over tasks.
Now let us choose
$K^2=\mathcal{O}(\epsilon_{\rm TF}^{-2})$ large enough so that 
\begin{equation}\label{eq:sm-truncation-choice}
    \E_w\bigl[U\mathbf1_{\mathcal E_K^c}\bigr]
    \leq\frac{\epsilon_{\rm TF}^2}{2},
    \qquad\text{for }U\in\{U_{\rm pred},U_{\rm obj},U_{\rm read}\}.
\end{equation}
Since $\gamma\leq\epsilon_{\rm TF}/2$, decomposing
expectations over $\mathcal E_K$ and $\mathcal E_K^c$ establishes the
objective-gap and readout bounds of \cref{thm:sm-main}:
\begin{equation*}
    \E_w\Bigl[J_\lambda(\widehat\vbeta)
       -\min_{\norm{\vu}_\infty\leq R}J_\lambda(\vu)\Bigr]
    \leq\gamma+\frac{\epsilon_{\rm TF}^2}2\leq\epsilon_{\rm TF},
    \qquad
    \E_w\bigl(\widehat y_{\rm TF}-\vpsi(\vX_{N+1})^\top\widehat\vbeta\bigr)^2
    \leq\gamma^2+\frac{\epsilon_{\rm TF}^2}2\leq\epsilon_{\rm TF}^2.
\end{equation*}

\paragraph{Prediction error on \texorpdfstring{$\mathcal E\cap\mathcal E_K$}{the good event}.}
Since $\norm{\vbeta_w^\star}_\infty\leq\norm{\vbeta_w^\star}_2\leq R$ by \cref{ass:sm-sparse}, the objective-gap bound on $\mathcal E_K$ implies
$J_\lambda(\widehat\vbeta)\leq J_\lambda(\vbeta_w^\star)+\gamma$. On
$\mathcal E\cap\mathcal E_K$, \cref{lem:sm-prompt-event-main} then gives
\begin{equation*}
    \norm{\widehat\vbeta-\vbeta_w^\star}_2
    \leq C\Big(\frac{\sqrt{s}\lambda}{\alpha}+\frac\gamma\lambda\Big).
\end{equation*}
Decomposing the prediction error as
\begin{equation}
    \begin{aligned}
        \widehat y_{\rm TF}-f_w(\vX_{N+1})
        &=\underbrace{\bigl(\widehat y_{\rm TF}-\vpsi(\vX_{N+1})^\top\widehat\vbeta\bigr)}_{\text{(I)}}
          +\underbrace{\vpsi(\vX_{N+1})^\top(\widehat\vbeta-\vbeta_w^\star)}_{\text{(II)}}\\
        &\quad+\underbrace{\bigl(\vpsi(\vX_{N+1})^\top\vbeta_w^\star-f_w(\vX_{N+1})\bigr)}_{\text{(III)}},
    \end{aligned}
\end{equation}
we have $|\text{(I)}|\leq\gamma$,
$|\text{(II)}|\leq\norm{\vpsi(\vX_{N+1})}_2\norm{\widehat\vbeta-\vbeta_w^\star}_2$
by Cauchy--Schwarz, and $|\text{(III)}|\leq\epsilon$ by
\cref{eq:sm-sparse-approximation} on $\mathcal E\cap\mathcal E_K$. Combining
these bounds, squaring, and taking expectations using
$\E_w\norm{\vpsi(\vX_{N+1})}_2^2\leq B^2$ (\cref{ass:sm-bounded}) yields
\begin{equation*}
    \E_w\Big[
       \bigl(\widehat y_{\rm TF}-f_w(\vX_{N+1})\bigr)^2
       \mathbf1_{\mathcal E\cap\mathcal E_K}\Big]
    \leq C\epsilon^2+
       \frac{CB^2s\lambda^2}{\alpha^2}
       +C\gamma^2\Big(1+\frac{B^2}{\lambda^2}\Big).
\end{equation*}
By \cref{ass:sm-rsc,ass:sm-bounded} and \cref{eq:sm-main-transformer-tuning},
\begin{equation*}
    s\lambda^2
    \leq2C^2(B^2+\alpha)\Big(\frac{B^2}{2\alpha}\epsilon^2
       +\frac{\sigma^2s\log(PN)}{N}\Big).
\end{equation*}
Since $\gamma^2(1+B^2/\lambda^2)\leq c_1^2\epsilon_{\rm TF}^2$, taking $c_1$ sufficiently small yields
\begin{equation}\label{eq:sm-general-score-rate}
    \E_w\Big[
       \bigl(\widehat y_{\rm TF}-f_w(\vX_{N+1})\bigr)^2
       \mathbf1_{\mathcal E\cap\mathcal E_K}\Big]
    \leq C\Big(\epsilon^2+
       \frac{\sigma^2s\log(PN)}{N}\Big)
       +\frac{\epsilon_{\rm TF}^2}{2}.
\end{equation}

\paragraph{Tail error.}
On the complement $(\mathcal E\cap\mathcal E_K)^c=\mathcal E^c\cup\mathcal E_K^c$,
we bound the contribution of each event separately. For $\mathcal E_K^c$,
\cref{eq:sm-global-prediction-bound,eq:sm-truncation-choice} give
\begin{equation*}
    \E_w\Big[
       \bigl(\widehat y_{\rm TF}-f_w(\vX_{N+1})\bigr)^2
       \mathbf1_{\mathcal E_K^c}\Big]
    \leq\E_w\bigl[U_{\rm pred}\mathbf1_{\mathcal E_K^c}\bigr]
    \leq\frac{\epsilon_{\rm TF}^2}{2}.
\end{equation*}
Meanwhile, on $\mathcal E^c$, applying Cauchy--Schwarz to
\cref{eq:sm-global-prediction-bound} with $\P_w(\mathcal E^c)\leq C(PN)^{-4}$ yields
\begin{equation*}
    \E_w\Big[
       \bigl(\widehat y_{\rm TF}-f_w(\vX_{N+1})\bigr)^2
       \mathbf1_{\mathcal E^c}\Big]
    \leq2R^2(\sqrt P+1)^2B^2\P_w(\mathcal E^c)^{1/2}
       +2\epsilon^2\P_w(\mathcal E^c)
    \leq\frac{C}{PN^2}
       +\frac{C\epsilon^2}{(PN)^4}.
\end{equation*}
Now, summing \cref{eq:sm-general-score-rate} with both tail bounds proves \cref{eq:sm-general-rate}:
\begin{equation*}
    \E_w\bigl(\widehat y_{\rm TF}-f_w(\vX_{N+1})\bigr)^2
    \leq C\Big(\epsilon^2+
       \frac{\sigma^2s\log(PN)}{N}
       +\frac{1}{PN^2}\Big)
       +\epsilon_{\rm TF}^2.
\end{equation*}

\paragraph{Transformer configuration.}
\Cref{eq:sm-main-transformer-tuning} gives
$\gamma^{-1}=\mathcal{O}(\epsilon_{\rm TF}^{-1})$. Combined with
$K^2=\mathcal{O}(\epsilon_{\rm TF}^{-2})$, taking the configuration in
\cref{lem:sm-transformer-lasso-main} yields a total number of layers of
\begin{equation*}
    L+1=1+\left\lceil\frac{(1+K^2)R^2P}{\gamma}\right\rceil
       =\mathcal{O}(\epsilon_{\rm TF}^{-3}),
\end{equation*}
And each layer contains two softmax attention heads with head dimension $P+1$, followed by an MLP block of width $4P+1$.

\section{Experiments}

The goal of our experiments is to demonstrate the effectiveness of the signature transform as a path representation for ICL with nonlinear path-dependent functionals. 
We will first examine sparse recovery of signature features with transformers, then evaluate
signature-based ICL on three experiments: regression on controlled differential
equations, optimal execution, and options with  path-dependent payoff using market data.

We will compare three path-embedding methods:
\emph{TF-Raw-Flat} applies a learnable map directly to the vectorized raw path observations; \emph{TF-Raw-Seq} processes the raw path through a sequence encoder to capture sequential structure; and \emph{TF-Sig} directly embeds the truncated signature features (omitting the constant zeroth-order term).
Within each experiment, the resulting path embeddings are then fed into a transformer, which is trained separately, to perform in-context prediction.
All results are aggregated over three independent runs.

\subsection{In-Context Sparse Recovery of Signature Features}
\label{sec:sparse-signature-recovery}

We first test whether \emph{TF-Sig} can recover the active signature
coordinates of an unknown sparse linear functional in context.

Let $\vZ$ be a standard five-dimensional Brownian motion over the time interval $[0,1]$, and define each input path $\vX$ as the piecewise linear interpolation of $17$ uniform observations of $\vZ$ at $t_j=j/16$ ($j=0,\ldots,16$).
We compute the normalized signature up to level three to get $\psi(\vX)\in\R^{155}$.
For each task, the ground-truth coefficient $\vbeta^\star\in\R^{155}$ has $16$ uniformly sampled active coordinates, with independent random signs and magnitudes drawn uniformly from $[0.7,1.3]/4$. The label is given by $Y=\vpsi(\vX)^\top\vbeta^\star$.

The path encoder $\vE_{\mathrm{Sig}}$ then maps $\vpsi(\vX)$ into $\R^{1152}$ via a linear projection followed by LayerNorm. Given learnable role embeddings $\vr_0,\vr_c,\vr_\beta\in\R^{1152}$, we construct the prompt sequence
\begin{equation}
\vz_0=\vr_0,\qquad
\vz_i=\mW_{\mathrm{in}}[\vE_{\mathrm{Sig}}(\vX_i);Y_i]+\vr_c,
\quad i=1,\ldots,N,\qquad
\vz_{N+1}=\vr_\beta.
\end{equation}
Here $\vz_{N+1}$ is a virtual readout token.
A $12$-layer, eight-head causal transformer then processes $(\vz_0,\ldots,\vz_{N+1})$ and a linear head maps the terminal state of $\vz_{N+1}$ directly to the coefficient estimate $\widehat\vbeta\in\R^{155}$.

We train \emph{TF-Sig} for $10^6$ steps across context lengths $N\in\{32,64,128\}$ by minimizing the prediction MSE together with an explicit coefficient loss $0.1\norm{\widehat\vbeta-\vbeta^\star}_2^2$, thereby providing direct supervision on $\vbeta^\star$ during training. As a baseline, we compare against \emph{Lasso-Sig} fit directly on $\vpsi(\vX)$, tuning its $\ell_1$ penalty via cross-validation.

\begin{table}[t]
    \centering
    \small
    \caption{Support recovery $F1$ scores, where the estimated support is defined by $|\widehat\beta_j|>0.05$. Contexts $N\in\{256,512\}$ extrapolate beyond \emph{TF-Sig}'s maximum training length ($N=128$).}
    \label{tab:sparse-signature-recovery}
    \begin{tabular}{lccccc}
        \toprule
        Model & $N=32$ & $N=64$ & $N=128$ & $N=256$ & $N=512$ \\
        \midrule
        Lasso-Sig & $0.403$ & $0.997$ & $1.000$ & $1.000$ & $1.000$ \\
        TF-Sig    & $0.404$ & $0.725$ & $0.988$ & $0.999$ & $1.000$ \\
        \bottomrule
    \end{tabular}
\end{table}

\paragraph{Results.}
As shown in \cref{tab:sparse-signature-recovery}, increasing the context length from $N=32$ to $N=128$ raises \emph{TF-Sig}'s support recovery F1 from $0.404$ to $0.988$. Moreover, support recovery generalizes smoothly to unseen context lengths, reaching $0.999$ at $N=256$ and $1.000$ at $N=512$. Because the ground-truth functional is strictly linear in $\vpsi(\vX)$, \emph{Lasso-Sig} is well-specified and recovers the exact support on every test task from $N=128$ onward. 

Overall, these results confirm that {a transformer with signature features can effectively simulate in-context Lasso for sparse feature recovery}.

\subsection{Robustness of the Signature to Path Observation Grids}
\label{sec:synthetic-cde-experiment}

\paragraph{Task setup.}
We construct this example to evaluate ICL on nonlinear path-dependent
functions that admit an exact signature representation.

Let $\vZ$ be a stationary five-dimensional OU process
\begin{equation}
    \dif\vZ_t=-\frac12\vZ_t\,\dif t+\mL\,\dif\vW_t,
    \qquad (\mL\mL^\top)_{rs}=0.6^{|r-s|},
    \quad 1\leq r,s\leq5,
\end{equation}
where $\vZ_0\sim\mathcal N(\vzero_5,\mL\mL^\top)$.
For $j=1,\ldots,N+1$, define the observed path on the uniform grid by
\begin{equation*}
    \vX_j(t_\ell)
    =\frac{\vZ_{j-1+t_\ell}-\vZ_{j-1}}{\sqrt{2(1-e^{-1/2})}},
    \qquad t_\ell=\frac\ell{32},\quad \ell=0,\ldots,32,
\end{equation*}
and interpolate linearly between grid points.
A task parameter $\vtheta\sim\mathcal N(0,\mI_3)$ then determines the controlled
differential equation driven by $\vX$:
\begin{equation}
    \dif \vu_t=\sum_{i=1}^5\mA_i(\vtheta)\vu_t\,\dif \vX^{(i)}(t),
    \qquad \vu_0=(0,0,0,0,1)^\top,\qquad 0\leq t\leq 1.
\end{equation}
Here $\mA_i(\vtheta)\in\R^{5\times5}$ is supported on its first superdiagonal:
\begin{equation*}
    [\mA_i(\vtheta)]_{r,r+1}
    =(\mC_i\vtheta)_r,\qquad r=1,\ldots,4,
\end{equation*}
with all other entries equal to zero, and $\mC_1,\ldots,\mC_5\in\R^{4\times 3}$ fixed matrices.
The observed label is given by $Y = \tanh(u_1^{(1)}) + \xi$, where $u_1^{(1)}$ denotes the first coordinate of $\vu_1$ and $\xi\sim\mathcal N(0,0.1^2)$ is independent Gaussian noise.
Because the matrices $\mA_i(\vtheta)$ are supported solely on the first superdiagonal, a product of these matrices has a nonzero $(1,5)$ entry if and only if it consists of exactly four factors. The signature expansion for linear controlled differential equations \citep[Section~4]{lyons2014rough} therefore reduces to the fourth signature level:
\begin{equation*}
\begin{aligned}
    u_1^{(1)}
    =\sum_{i_1,\ldots,i_4=1}^5
       [\mA_{i_4}(\vtheta)\cdots\mA_{i_1}(\vtheta)]_{1,5}
       [\Sig^{(4)}(\vX)]_{i_1,\ldots,i_4}.
\end{aligned}
\end{equation*}
Consequently, the fourth-order truncated signature retains all path information required to determine the label.

\paragraph{Path and prompt encoding.}
\emph{TF-Raw-Flat}, \emph{TF-Raw-Seq}, and \emph{TF-Sig} have different path encoders,
$\vE_{\mathrm{Flat}},\vE_{\mathrm{Seq}},\vE_{\mathrm{Sig}}$, respectively,
with embedding dimension 128.
\emph{TF-Raw-Flat} linearly projects the $33\times5=165$ flattened path coordinates to
$\vE_{\mathrm{Flat}}(\vX)$.
\emph{TF-Raw-Seq} encodes each path's 32 points with a two-layer gated recurrent unit (GRU), then linearly projects its final hidden state to $\vE_{\mathrm{Seq}}(\vX)$.
\emph{TF-Sig} computes $\vE_{\mathrm{Sig}}(\vX)$ by linearly projecting the normalized fourth-order truncated signature $\vpsi(\vX)\in\R^{780}$.
Given a learnable affine label encoder $\vphi_y$, fixed sinusoidal embeddings $\vp_j$ indexed by example \citep{vaswani2017attention}, and learnable path/label type embeddings $\vr_x,\vr_y$, the input tokens are defined as
\begin{equation}
    \vz_{2j-2}=\vE(\vX_j)+\vp_j+\vr_x,\quad 1\leq j\leq 33,
    \qquad
    \vz_{2j-1}=\vphi_y(y_j)+\vp_j+\vr_y,\quad 1\leq j\leq 32.
\end{equation}
The prompt
$(\vz_0,\ldots,\vz_{64})
\in\R^{65\times128}$ is then fed into a four-layer global causal transformer with 8 attention heads.
During training, a shared linear readout head predicts each label $Y_j$ from the preceding $j-1$ demonstrations, and the model is optimized by averaging the prediction losses across all positions.

\begin{table}[t]
    \centering
    \small
    \caption{MSE across different observation grids.
    Interval denotes the mean spacing between observations.
    Because TF-Raw-Flat requires a fixed input of $33$ grid points for each path, paths with $9$ or $17$ observations are linearly interpolated onto the $33$-point training grid (ID).}
    \label{tab:synthetic-cde-ood}
    \begin{tabular}{lrccc}
        \toprule
        Grid & Interval & TF-Raw-Flat & TF-Raw-Seq & TF-Sig \\
        \midrule
        Uniform (ID) & $1/32$
            & $0.729_{\pm 0.012}$ & $0.373_{\pm 0.022}$ & $\mathbf{0.160_{\pm 0.015}}$ \\
        Uniform & $1/16$
            & $0.738_{\pm 0.012}$ & $0.513_{\pm 0.021}$ & $\mathbf{0.340_{\pm 0.009}}$ \\
        Uniform & $1/8$
            & $0.749_{\pm 0.004}$ & $0.708_{\pm 0.023}$ & $\mathbf{0.610_{\pm 0.023}}$ \\
        Nonuniform & $1/32$
            & $0.757_{\pm 0.016}$ & $0.426_{\pm 0.012}$ & $\mathbf{0.277_{\pm 0.011}}$ \\
        Nonuniform & $1/16$
            & $0.732_{\pm 0.009}$ & $0.587_{\pm 0.009}$ & $\mathbf{0.460_{\pm 0.010}}$ \\
        \bottomrule
    \end{tabular}
\end{table}

\paragraph{Results.}
At test time, we predict the query label $Y_{33}$ conditioned on all $N=32$ demonstrations and report the mean squared error (MSE) and $R^2$ against the ground-truth targets.
Beyond the 33-point uniform grid used in training, at test time we vary both the number and the placement of observation times for the demonstration and query paths to assess the ICL robustness.
As shown in \cref{tab:synthetic-cde-ood}, {signature-based ICL is robust with respect to changes in the path observation grids.} On the in-distribution (ID) training grid, \emph{TF-Sig} attains an $R^2$ of $0.85$, substantially outperforming \emph{TF-Raw-Seq} ($0.64$) and \emph{TF-Raw-Flat} ($0.30$). 

Across all evaluated uniform and nonuniform test grids, \emph{TF-Sig} consistently achieves the lowest MSE. Although coarser discretizations and irregular sampling increase the error of both \emph{TF-Raw-Seq} and \emph{TF-Sig}, \emph{TF-Sig} maintains a clear advantage across every regime.

\subsection{Optimal Execution: Adaptation to Unseen Market Environments}
\label{sec:execution-cost-icl}

\paragraph{Task setup.}
We consider a financial optimal-execution task with unknown transient
price impact, using kernel families adapted from
\citet{meng2026optimalexecution}. Given a trading-rate schedule $u=(u_t)_{t\in[0,T]}$, let $Q_t=\int_0^t u_s\,\dif s$ denote the cumulative traded volume, and define the augmented state path $\vX=(t,Q_t,u_t)_{t\in[0,T]}$. A latent task parameter $\vtheta=(\lambda,\delta,\eta)$ governs the transient price impact $I_{\vtheta,u}(t)$ and the resulting per-unit execution cost $f_\vtheta(\vX)$:
\begin{equation}
    \begin{aligned}
        I_{\vtheta,u}(t)
        &=\lambda\int_0^t G_\eta(t-s)
            \operatorname{sgn}(u_s)\abs{u_s}^{\delta}\,\dif s,\\
        f_\vtheta(\vX)
        &=\frac{1}{\abs{Q_T}}\int_0^T u_tI_{\vtheta,u}(t)\,\dif t,
        \qquad \vX=((t,Q_t,u_t))_{t\in[0,T]}.
    \end{aligned}
\end{equation}
Within each prompt, the impact parameter $\vtheta$ (and hence the impact operator) is fixed, while the execution schedules $u$ are generated by transforming Gaussian process draws into positive trading-rate profiles. Each demonstration label is given by $Y=f_\vtheta(\vX)+\xi$, where $\xi$ is an independent Gaussian noise.

The models are trained with $N\in\{1,2,4,8,16\}$ on the normalized
exponential kernel
$G_\eta(r)=\frac{\eta e^{-\eta r}}{1-e^{-\eta}}$.
At test time, they are evaluated on in-distribution exponential kernels as well as two unseen normalized power-law families: the non-singular kernel $G_\eta(r)=\frac{(1+r)^{-\eta}}{\!\int_0^1(1+s)^{-\eta}\,\dif s}$ and the singular kernel $G_\eta(r)=(1-\eta)r^{-\eta}$ for $0<\eta<1$.

\paragraph{Path and prompt encoding.}
Each execution path $\vX$ is discretized as $\vx_k=(t_k,Q_{t_k},u_{t_k})\in\R^3$ ($k=0,\ldots,100$) and mapped into $\R^{64}$ via one of three path encoders.
Specifically, \emph{TF-Raw-Flat} applies LayerNorm followed by a linear projection to the $303$ flattened coordinates to produce $\vE_{\mathrm{Flat}}(\vX)$, whereas \emph{TF-Raw-Seq} passes the $101$-step sequence through a two-layer local causal transformer and extracts the terminal state as $\vE_{\mathrm{Seq}}(\vX)$. For \emph{TF-Sig}, we first lift $(\vx_k)_{k=0}^{100}$ to the lead--lag path $\operatorname{LeadLag}(\vX)=(\vL_j)_{j=0}^{200}$ \citep{gyurko2014extracting}:
\begin{equation}
\vL_0=(\vx_0,\vx_0),\qquad
\vL_{2k-1}=(\vx_k,\vx_{k-1}),\qquad
\vL_{2k}=(\vx_k,\vx_k),\quad k=1,\ldots,100.
\end{equation}
From this fixed lead--lag lifting, we compute the normalized signature features $\vpsi(\vX)\in\R^{258}$ up to level three and pass them through a GELU MLP to obtain $\vE_{\mathrm{Sig}}(\vX)$.
For the prompt format, we concatenate each path embedding with its scalar label (using $0$ for the query), apply a shared linear projection $\mW_{\mathrm{in}}:\R^{65}\to\R^{64}$, and add learnable beginning-of-sequence, context, and query role embeddings $\vr_0,\vr_c,\vr_q\in\R^{64}$:
\begin{equation}\label{eq:experiment-prompt-encoding}
\vz_0=\vr_0,\qquad
\vz_i=\mW_{\mathrm{in}}[\vE(\vX_i);y_i]+\vr_c,\quad i=1,\ldots,N,
\qquad
\vz_{N+1}=\mW_{\mathrm{in}}[\vE(\vX_{N+1});0]+\vr_q.
\end{equation}
The resulting sequence $(\vz_0,\ldots,\vz_{N+1})\in\R^{(N+2)\times64}$ is fed into a two-layer transformer with 4 attention heads.

As a training-free baseline, \emph{Lasso-Sig} fits a Lasso regression to the same signature features in each prompt, where the $\ell_1$-penalty is selected via cross-validation over demonstration
labels.

\begin{table}[t]
    \centering
    \caption{In-distribution (ID) test $R^2$ on exponential kernels. The $N=32$ column evaluates context-length extrapolation beyond the maximum training length ($N=16$).}
    \label{tab:execution-cost-id-results}
    \resizebox{\textwidth}{!}{%
    \begin{tabular}{l|r|rrrrrr}
        \toprule
        & & \multicolumn{6}{c}{$R^2\ \uparrow$} \\[-0.25em]
        Model & \# Params. & $N=1$ & $N=2$ & $N=4$ & $N=8$ & $N=16$ & $N=32$ \\
        \midrule
        Lasso-Sig & N.A.
            & $\detval{0.184}$ & $\detval{0.381}$ & $\detval{0.532}$
            & $\detval{0.724}$ & $\detval{0.908}$ & $\detval{0.930}$ \\
        TF-Raw-Flat & 0.125M
            & $0.931_{\pm 0.001}$ & $0.964_{\pm 0.001}$
            & $0.980_{\pm 0.001}$ & $0.987_{\pm 0.001}$
            & $0.990_{\pm 0.001}$ & $0.992_{\pm 0.001}$ \\
        TF-Raw-Seq & 0.209M
            & $0.934_{\pm 0.001}$ & $0.967_{\pm 0.000}$
            & $0.982_{\pm 0.000}$ & $0.988_{\pm 0.000}$
            & $0.991_{\pm 0.000}$ & $0.993_{\pm 0.000}$ \\
        TF-Sig & 0.126M
            & $\mathbf{0.934_{\pm 0.001}}$ & $\mathbf{0.968_{\pm 0.000}}$
            & $\mathbf{0.983_{\pm 0.000}}$ & $\mathbf{0.989_{\pm 0.000}}$
            & $\mathbf{0.993_{\pm 0.000}}$ & $\mathbf{0.994_{\pm 0.000}}$ \\
        \bottomrule
    \end{tabular}
    }
\end{table}

\begin{table}[t]
    \centering
    \caption{Out-of-distribution (OOD) test $R^2$ on unseen power-law kernels.}
    \label{tab:execution-cost-ood-power-results}
    \begin{tabular}{l|rrrr}
        \toprule
        & \multicolumn{4}{c}{$R^2\ \uparrow$} \\[-0.25em]
        & \multicolumn{2}{c}{Non-singular} & \multicolumn{2}{c}{Singular} \\
        \cmidrule(lr){2-3}\cmidrule(lr){4-5}
        Model & $N=16$ & $N=32$ & $N=16$ & $N=32$ \\
        \midrule
        Lasso-Sig & $\detval{0.839}$ & $\detval{0.911}$
            & $\detval{0.855}$ & $\detval{0.938}$ \\
        TF-Raw-Flat & $0.938_{\pm 0.004}$ & $0.946_{\pm 0.003}$
            & $0.982_{\pm 0.001}$ & $0.986_{\pm 0.002}$ \\
        TF-Raw-Seq  & $0.943_{\pm 0.002}$ & $0.951_{\pm 0.001}$
            & $0.984_{\pm 0.001}$ & $0.988_{\pm 0.001}$ \\
        TF-Sig & $\mathbf{0.954_{\pm 0.005}}$ & $\mathbf{0.963_{\pm 0.005}}$
            & $\mathbf{0.986_{\pm 0.001}}$ & $\mathbf{0.990_{\pm 0.001}}$ \\
        \bottomrule
    \end{tabular}
\end{table}

\paragraph{Results.}
The results are reported in
\cref{tab:execution-cost-id-results,tab:execution-cost-ood-power-results}.
On the in-distribution exponential kernels (\cref{tab:execution-cost-id-results}), \emph{TF-Sig} achieves the highest $R^2$ at every context length and extrapolates smoothly to $N=32$. Under OOD shift to the unseen non-singular and singular power-law kernels, \emph{TF-Sig} attains $R^2$ values of $0.963$ and $0.990$ at $N=32$, outperforming \emph{TF-Raw-Seq} in every paired run while requiring $39.8\%$ fewer parameters.

Note that the training-free baseline {Lasso-Sig} performs substantially worse than {TF-Sig} with the same signature features, indicating that {the transformer provides essential expressivity when the signature truncation level is insufficient to fully linearize the target functional, while leveraging cross-task
priors from pre-training that go beyond single-task Lasso fitting.}

\subsection{Options with Path-Dependent Payoffs}
\label{sec:financial-payoff-icl}

\paragraph{Task setup.}
We evaluate ICL on Asian and variance option payoffs
\citep{kemna1990pricing,carr2009volatility,kellerressel2013asymptotic},
a classic family of path-dependent derivatives previously studied through the signature \citep{perezarribas2018derivatives}.

Each path $\vX=(\tau_t,x_t)_{t=0}^{31}$ comprises $32$ observations over a five-minute window, where $\tau_t=t/31$ and $x_t=10^4\log(S_t/S_0)$ denotes the cumulative log return.
For $\vX$, we compute the realized average and variance:
\begin{equation}
A(\vX)=\frac{1}{32}\sum_{t=0}^{31}\exp(x_t/10^4),
\qquad
V(\vX)=\frac{1}{31}\sum_{t=1}^{31}(x_t-x_{t-1})^2.
\end{equation}
Each task is specified by a latent contract $(g,\omega,K)$ consisting of a payoff type $g\in\{A,V\}$, an option side $\omega\in\{+1,-1\}$ (call or put) and a strike $K$, yielding the terminal payoff
\begin{equation}
f(\vX)=c_g[\omega(g(\vX)-K)]_+,
\qquad c_A=10^4,\quad c_V=1.
\end{equation}
The contract is fixed within each prompt but remains unobserved; because an identical price trajectory yields distinct payoffs under different contracts, the model must infer the underlying payoff structure from the context demonstrations.

Price paths are constructed from transaction records in Binance Public Data \citep{binancepublicdata}. The training split comprises $4{,}000$ paths from Bitcoin (BTC), Ether (ETH), BNB, Solana (SOL), XRP, and Cardano (ADA) traded against USDT in 2023, whereas validation and testing use $2{,}000$ and $4{,}000$ paths, respectively, drawn from two disjoint pairs of unseen assets in 2024 and 2025. Training prompts sample strikes from a continuous range. Within every prompt, the demonstration and query paths are distinct, and we use contexts of at most $N=16$ demonstrations for training.

\paragraph{Path and prompt encoding.}
\emph{TF-Raw-Flat} and \emph{TF-Raw-Seq} adapt the flattened-coordinate and two-layer local causal transformer encoders of \cref{sec:execution-cost-icl} to map each $32$-step path $\vX$ into $\R^{128}$.
For \emph{TF-Sig}, we construct a partial lead--lag path that duplicates only the price channel \citep{gyurko2014extracting}:
\begin{equation}
    \vL_0=(\tau_0,x_0,x_0),\quad
    \vL_{2k-1}=(\tau_k,x_k,x_{k-1}),\quad
    \vL_{2k}=(\tau_k,x_k,x_k),\quad k=1,\ldots,31.
\end{equation}
From the resulting $63$-point trajectory $\operatorname{LeadLag}(\vX)=(\vL_j)_{j=0}^{62}$ in $\R^3$, we extract the sixth-order truncated signature $\vpsi(\vX)\in\R^{1092}$ and apply a learnable linear projection followed by LayerNorm to obtain $\vE_{\mathrm{Sig}}(\vX)\in\R^{128}$.
Finally, the prompt sequence $(\vz_0,\ldots,\vz_{N+1})\in\R^{(N+2)\times128}$, ordered as in \cref{eq:experiment-prompt-encoding}, is processed by a two-layer transformer with eight attention heads.

\begin{table}[t]
    \centering
    \caption{Test MSE, averaged over contracts on held-out
    2025 assets and unseen strikes.}
    \label{tab:finance-functional-results}
    \begin{tabular}{l|r|rrrr}
        \toprule
        & & \multicolumn{4}{c}{Normalized MSE $\downarrow$} \\[-0.25em]
        Model & \# Params. & $N=4$ & $N=8$ & $N=16$ & $N=32$ \\
        \midrule
        Lasso-Sig & N.A. & $\detval{0.974}$ & $\detval{0.819}$
            & $\detval{0.653}$ & $\detval{0.467}$ \\
        TF-Raw-Flat & 0.55M & $0.269_{\pm 0.005}$ & $0.158_{\pm 0.005}$
            & $0.112_{\pm 0.004}$ & $0.103_{\pm 0.005}$ \\
        TF-Raw-Seq  & 0.83M & $0.239_{\pm 0.012}$ & $0.153_{\pm 0.005}$
            & $0.112_{\pm 0.006}$ & $0.105_{\pm 0.008}$ \\
        TF-Sig & 0.55M
            & $\mathbf{0.160_{\pm 0.018}}$ & $\mathbf{0.076_{\pm 0.004}}$
            & $\mathbf{0.033_{\pm 0.002}}$ & $\mathbf{0.023_{\pm 0.003}}$ \\
        \bottomrule
    \end{tabular}
\end{table}

\begin{table}[t]
    \centering
    \caption{Test MSE in different
    payoff families.}
    \label{tab:finance-family-results}
    \begin{tabular}{l|rrrr}
        \toprule
        & \multicolumn{4}{c}{Normalized MSE $\downarrow$} \\[-0.25em]
        & \multicolumn{2}{c}{Asian} & \multicolumn{2}{c}{Variance} \\
        \cmidrule(lr){2-3}\cmidrule(lr){4-5}
        Model & $N=16$ & $N=32$ & $N=16$ & $N=32$ \\
        \midrule
        Lasso-Sig & $\detval{0.205}$ & $\detval{0.111}$
            & $\detval{1.101}$ & $\detval{0.824}$ \\
        TF-Raw-Flat & $\mathbf{0.007_{\pm 0.001}}$ & $\mathbf{0.006_{\pm 0.001}}$
            & $0.218_{\pm 0.008}$ & $0.201_{\pm 0.010}$ \\
        TF-Raw-Seq  & $0.009_{\pm 0.001}$ & $0.008_{\pm 0.001}$
            & $0.214_{\pm 0.012}$ & $0.203_{\pm 0.017}$ \\
        TF-Sig & $\mathbf{0.007_{\pm 0.002}}$ & $\mathbf{0.006_{\pm 0.002}}$
            & $\mathbf{0.060_{\pm 0.002}}$ & $\mathbf{0.040_{\pm 0.005}}$ \\
        \bottomrule
    \end{tabular}
\end{table}

\paragraph{Results.}
As reported in \cref{tab:finance-functional-results,tab:finance-family-results}, \emph{TF-Sig} achieves the lowest MSE across all context lengths. At $N=16$, it reaches an MSE of $0.033$—a $70.2\%$ reduction relative to the strongest raw-data baseline—while requiring $33.4\%$ fewer parameters than \emph{TF-Raw-Seq}. At the extrapolated context length $N=32$, \emph{TF-Sig} further lowers the MSE to $0.023$ ($77.6\%$ below the best raw baseline), whereas \emph{Lasso-Sig} attains only $0.467$ on the exact same signature features, highlighting the role of the transformer's nonlinear in-context processing.

{Overall, signature features enable accurate in-context learning of nonlinear, path-dependent financial payoffs with a compact transformer.}

Decomposing the performance by payoff family (\cref{tab:finance-family-results}) reveals that the gain is primarily driven by variance options: at $N=32$, \emph{TF-Sig} achieves an MSE of $0.040$ versus $0.201$ and $0.203$ for the raw baselines, whereas all three transformer models perform comparably on Asian options. This gap admits a direct explanation through the signature representation. In fact, the second-level signature of the lead-lag path $\vL=\operatorname{LeadLag}(\vX)$ satisfies the exact discrete quadratic-variation identity \citep{chevyrev2025primer}:
\begin{equation}\label{eq:payoff-leadlag-variance}
\frac{1}{31}[\Sig^{2,3}(\vL)-\Sig^{3,2}(\vL)]
=\frac{1}{31}\sum_{k=1}^{31}(\Delta x_k)^2
=V(\vX).
\end{equation}
Consequently, the realized variance $V(\vX)$ is recovered exactly as a linear combination of two second-level signature coordinates, substantially enhancing the transformer's ICL capability.

\section{Conclusion}

We integrate path signatures with transformers to enable universal in-context learning of path-dependent functionals through explicit, interpretable path features. 
Using the signature transform as a
fixed path embedding, our theoretical construction emulates in-context signature Lasso and establishes finite-sample prediction guarantees under
mild conditions, allowing dependence among demonstrations and the query. 
Experiments demonstrate that signature-based transformers match or outperform vanilla transformer baselines with substantially fewer trainable parameters, exhibiting strong expressivity for path quantities, robustness across observation grids, and effective adaptation to unseen market regimes. 
Together, these results establish a principled foundation for path-dependent universal ICL that combines interpretable features with flexible in-context adaptation.

\bibliographystyle{plainnat}
\bibliography{ref}

\newpage
\appendix
\crefalias{section}{appendix}
\crefalias{subsection}{appendix}
\crefalias{subsubsection}{appendix}

\section{Signature Approximation and Statistical Analysis}
\label{app:sm-proofs}

\subsection{Signature Approximation over a Function Class}
\label{app:proof_sig_approx}
\begin{proof}[Proof of \cref{lem:sig-ua}]
For a single function, this is the signature approximation theorem
\citep[Theorem~3.1]{levin2013learning}: every continuous functional on
$\mathcal K$ is approximated uniformly by linear functionals of a finite
truncation.

For any $\epsilon>0$, by total boundedness, $\mathcal F$ admits a finite
$\epsilon/2$-net $f_1,\ldots,f_{N}$ in the supremum norm.
For each $f_j$, the fixed-function result provides a truncation level
$M_j$ and coefficients $\vbeta_j$ that approximate $f_j$ with error at
most $\epsilon/2$. Set $M:=\max_{1\leq j\leq N}M_j$.
For each $j$, append zero coefficients for the signature coordinates at
levels $M_j+1,\ldots,M$, so that $\vbeta_j\in\R^{P_M}$.
This leaves the approximation unchanged and allows every $f_j$ to be
approximated using the same truncation level $M$.
For any $f\in\mathcal F$, choose $j$ with
$\norm{f-f_j}_\infty\leq\epsilon/2$ and set $\vbeta_f:=\vbeta_j$.
By the triangle inequality,
\begin{equation*}
    \sup_{\vx\in\mathcal K}
    \abs{f(\vx)-\vbeta_f^\top\Sig_M(\widetilde{\vx})}
    \leq\norm{f-f_j}_\infty
       +\sup_{\vx\in\mathcal K}
          \abs{f_j(\vx)-\vbeta_j^\top\Sig_M(\widetilde{\vx})}
    \leq\epsilon.
\end{equation*}
The same truncation level $M$ therefore works for any $f\in\mathcal F$.
\end{proof}

\subsection{Technical Lemmas}
\label{app:proof_concentration}

This section collects the technical lemmas used to verify the Gram
concentration condition and establish the Lasso oracle inequality. 

Using a standard covering argument \citep{basu2015regularized}, the
following lemma reduces uniform quadratic deviations over the Lasso cone
to deviations along finitely many sparse directions.

\begin{lemma}[Quadratic deviations on the Lasso cone]
\label{lem:sm-restricted-net}
There is a finite set $\mathcal N\subset\R^P$ of unit vectors, with
$|\mathcal N|\leq(9P)^{2s}$, such that for every symmetric
$\mA\in\R^{P\times P}$ and every $S\subseteq\{1,\ldots,P\}$ with
$|S|\leq s$,
\begin{equation}\label{eq:sm-restricted-net}
    \sup_{\vv\in\mathcal C(S),\norm{\vv}_2\leq1}
       |\vv^\top\mA\vv|
    \leq72\max_{\vu\in\mathcal N}|\vu^\top\mA\vu|.
\end{equation}
Here $\mathcal C(S)$ is the cone defined in
\cref{eq:sm-population-covariance}.
\end{lemma}

\begin{proof}[Proof of \cref{lem:sm-restricted-net}]
For an index set $J\subseteq\{1,\ldots,P\}$, write $\vv_J$ for the subvector of a
vector $\vv$ indexed by $J$ and $\mA_{JJ}$ for the corresponding principal submatrix of $\mA$. Let $k=\min\{2s,P\}$. In each $k$-dimensional coordinate subspace, take a $1/4$-net of the unit sphere, and let $\mathcal N$ be the union of these nets, so that $|\mathcal N|\leq\binom Pk9^k\leq(9P)^{2s}$.
For any $|J|=k$, define $\vw\in\R^P$ as the unit vector supported on $J$ with $|\vw^\top\mA\vw|=\norm{\mA_{JJ}}_{\op}$.
By definition, we can choose $\vu\in\mathcal N$ supported on $J$ satisfying $\norm{\vw-\vu}_2\leq1/4$. Then
\begin{equation*}
|\vw^\top\mA\vw-\vu^\top\mA\vu|
\leq 2\norm{\mA_{JJ}}_{\op}\norm{\vw-\vu}_2
\leq\tfrac12\norm{\mA_{JJ}}_{\op}.
\end{equation*}
Since $|\vw^\top\mA\vw|=\norm{\mA_{JJ}}_{\op}$, it follows that
\begin{equation*}
    \norm{\mA_{JJ}}_{\op}
    \leq2\max_{\vu\in\mathcal N}|\vu^\top\mA\vu|.
\end{equation*}
Note that the same bound also holds for $|J|<k$ by trivially enlarging $J$ to a set of size $k$.
Now fix $\vv\in\mathcal C(S)$, sort its coordinates by decreasing
magnitude, and split them into consecutive blocks $J_1,J_2,\ldots$ of
size $s$, the last of which may be smaller. For $j\geq2$, we have
$\norm{\vv_{J_j}}_2\leq\norm{\vv_{J_{j-1}}}_1/\sqrt s$, while
$\norm{\vv}_1\leq5\sqrt s\norm{\vv}_2$ by the definition of $\mathcal C(S)$. Hence
$\sum_j\norm{\vv_{J_j}}_2\leq6\norm{\vv}_2$.
For any two blocks $J_j$ and $J_\ell$, the union $J=J_j\cup J_\ell$ has
at most $k$ indices, therefore
\begin{equation*}
    |\vv_{J_j}^\top\mA_{J_jJ_\ell}\vv_{J_\ell}|
    \leq\norm{\mA_{JJ}}_{\op}\cdot\norm{\vv_{J_j}}_2\norm{\vv_{J_\ell}}_2
    \leq 2\max_{\vu\in\mathcal N}|\vu^\top\mA\vu|\cdot \norm{\vv_{J_j}}_2\norm{\vv_{J_\ell}}_2.
\end{equation*}
Summing over all pairs of blocks, we obtain
\begin{equation*}
|\vv^\top\mA\vv|
\leq\sum_{j,\ell}|\vv_{J_j}^\top\mA_{J_jJ_\ell}\vv_{J_\ell}|
\leq2\max_{\vu\in\mathcal N}|\vu^\top\mA\vu|
\big(\sum_j\norm{\vv_{J_j}}_2\big)^2
\leq72\max_{\vu\in\mathcal N}|\vu^\top\mA\vu|\norm{\vv}_2^2.
\end{equation*}
\end{proof}

To control deviations along these sparse directions under dependent path
observations, we require a uniform concentration bound over a finite
family of functions. The following lemma combines a standard blocking and
coupling argument for $\beta$-mixing sequences
\citep{barrera2021generalization} with Bernstein's inequality
applied to independent block copies.
The next two lemmas are stated on an arbitrary probability space, with probability $\P$ and expectation $\E$.

\begin{lemma}[Bernstein inequality for mixing paths]
\label{lem:sm-blocking}
Let $(U_i)_{i\geq1}$ be a sequence of random variables in a standard Borel space. For
measurable functions $g_1,\ldots,g_J$, assume
$|g_j(U_i)|\leq A$ almost surely and $\E[g_j(U_i)^2]\leq B^4$
for any $i,j$, where $A,B>0$.
Writing $\mathcal F_n^U=\sigma(U_1,\ldots,U_n)$, define, for $k\geq1$,
\begin{equation}\label{eq:sm-beta-definition}
    \beta_U(k):=\sup_{n\geq1}\E\Big[
       \big\|\mathcal L((U_j)_{j\geq n+k}\mid\mathcal F_n^U)
       -\mathcal L((U_j)_{j\geq n+k})\big\|_{\TV}\Big].
\end{equation}
Here $\mathcal L$ denotes law
.
For integers $1\leq q\leq N$, put $m=\lfloor N/q\rfloor$.  Then, for
any $u>0$,
\begin{equation}\label{eq:sm-blocking}
\begin{split}
    \P\Big(
       \max_{1\leq j\leq J}
       \abs{\frac1N\sum_{i=1}^N\bigl(g_j(U_i)-\E g_j(U_i)\bigr)}>u
    \Big)
    \leq
    2J\exp\Big(-\frac{mu^2}{2B^4+4Au/3}\Big)+(N-q)\beta_U(q).
\end{split}
\end{equation}
\end{lemma}

\begin{proof}[Proof of \cref{lem:sm-blocking}]
Partition the indices into $q$ classes
$I_r=\{r,r+q,\ldots\}\cap\{1,\ldots,N\}$, $r=1,\ldots,q$,
each of size at least $m$.
By the coupling lemma of \citet[Lemma~2.10]{barrera2021generalization},
there are copies $U_i^\star$ with $\mathcal L(U_i^\star)=\mathcal L(U_i)$
that are independent within each class.
In addition, the coupling satisfies
$U_r^\star=U_r$ for $1\leq r\leq q$. And for $q<i\leq N$, $\P(U_i^\star\neq U_i)\leq\beta_U(q)$.
Therefore,
\begin{equation}\label{eq:sm-sequential-coupling}
    \P(\exists i\leq N:U_i^\star\neq U_i)
    \leq(N-q)\beta_U(q).
\end{equation}
Let $D_{j,r}=|I_r|^{-1}\sum_{i\in I_r}
\bigl(g_j(U_i^\star)-\E g_j(U_i)\bigr)$.
Within each class the summands are independent, so by Bernstein's
inequality \citep[Section~2.1.3]{wainwright2019high},
\begin{equation*}
    \E e^{tD_{j,r}}
    \leq\exp\Big(\frac{t^2B^4}{2m(1-2A|t|/(3m))}\Big),
    \qquad |t|<\frac{3m}{2A}.
\end{equation*}
By Jensen's inequality, the same bound holds for the weighted average
$\sum_r(|I_r|/N)D_{j,r}$, which equals the original centered mean on the event $\{U_i^\star=U_i,\ \forall i\}$.
We conclude the proof by applying Chernoff's bound \citep[Section~2.1.1]{wainwright2019high} to both tails, taking a union bound over $j$, and adding the coupling failure probability in \cref{eq:sm-sequential-coupling}.
\end{proof}

We now combine \cref{lem:sm-restricted-net,lem:sm-blocking} to bound
Gram deviations on the Lasso cone for unit-time segments of a geometrically
ergodic Markov process, allowing a nonstationary start. This bound is used
below to verify \cref{ass:sm-concentration} for the diffusion examples.

\begin{lemma}[Gram concentration  under geometric mixing]
\label{lem:sm-markov-gram}
Let $\vZ$ be a continuous time-homogeneous Markov process with
transition kernel $P_t$ and invariant probability measure $\pi$.
Suppose that, for a measurable function $V\geq1$ and constants $c,C>0$,
\begin{equation*}
    \norm{P_t(\vz,\cdot)-\pi}_{\TV}\leq CV(\vz)e^{-ct},
    \qquad \sup_{t\geq0}\E V(\vZ_t)\leq C.
\end{equation*}
For $\vX_i=(\vZ_{i-1+t})_{0\leq t\leq1}$, assume that, for some
$\rho\geq0$ and any $q\geq 2$,
$\sup_i\Big(\E\norm{\vpsi(\vX_i)}_2^q\Big)^{1/q}\leq Cq^\rho$.
For any $S\subseteq\{1,\ldots,P\}$ satisfying
$|S|\leq s$,
with probability at least $1-\delta$,
\begin{equation}\label{eq:sm-markov-gram}
\begin{split}
    \sup_{\vv\in\mathcal C(S),\norm{\vv}_2\leq1}
       |\vv^\top(\mG_N-\E[\mG_N])\vv|
    \leq C\Big(
       \sqrt{\frac{s\log(P/\delta)\log(N/\delta)}{N}}
       +\frac{s\log(P/\delta)[\log(N/\delta)]^{2\rho+1}}{N}
    \Big).
\end{split}
\end{equation}
The constant depends only on $\rho$ and the constants in the hypotheses.
\end{lemma}

\begin{proof}
The observations $(\vX_j)_{j\geq n+k}$ depend on the process from time
$n+k-1$ onward. By the Markov property, conditional on
$\vX_1,\ldots,\vX_n$, the state $\vZ_{n+k-1}$ has distribution
$P_{k-1}(\vZ_n,\cdot)$. Let $\mu_{n,k}$ denote its marginal
distribution and thus $\mu_{n,k}(\cdot)=\E[P_{k-1}(\vZ_n,\cdot)]$.
Therefore $\norm{\mu_{n,k}-\pi}_{\TV}\leq\E\norm{P_{k-1}(\vZ_n,\cdot)-\pi}_{\TV}$.
The conditional and marginal laws of $(\vX_j)_{j\geq n+k}$ are
obtained from these two state distributions through the same Markov
kernel. By the data-processing inequality,
\begin{equation*}
\begin{split}
    \beta_{\vX}(k)
    &\leq\sup_n\E\norm{P_{k-1}(\vZ_n,\cdot)-\mu_{n,k}}_{\TV}\\
    &\leq\sup_n\Big(
       \E\norm{P_{k-1}(\vZ_n,\cdot)-\pi}_{\TV}
       +\norm{\mu_{n,k}-\pi}_{\TV}\Big)\\
    &\leq2\sup_n\E\norm{P_{k-1}(\vZ_n,\cdot)-\pi}_{\TV}\\
    &\leq Ce^{-c(k-1)}\sup_n\E V(\vZ_n)\leq Ce^{-ck}.
\end{split}
\end{equation*}
Let $\mathcal N$ be the net in \cref{lem:sm-restricted-net} and set
$A=C[\log(2N/\delta)]^{2\rho}$.
By Markov's inequality with exponent $4\log(2N/\delta)$,
\begin{equation*}
    \sup_i\P(\norm{\vpsi(\vX_i)}_2^2>A)
    \leq\big(\frac{\delta}{2N}\big)^4.
\end{equation*}
Define $g_{\vu}(\vx):=\min\{(\vu^\top\vpsi(\vx))^2,A\}$ for
$\vu\in\mathcal N$. Since each $\vu$ is a unit vector,
$(\vu^\top\vpsi(\vX_i))^2\leq\norm{\vpsi(\vX_i)}_2^2$.
A union bound over $i$ shows that
$\max_{i\leq N}\norm{\vpsi(\vX_i)}_2^2\leq A$ with probability at
least $1-\delta/4$. On this event, truncation leaves all observed
values unchanged: $g_{\vu}(\vX_i)=(\vu^\top\vpsi(\vX_i))^2$
simultaneously for all $i\leq N$ and $\vu\in\mathcal N$.
The fourth-moment bound gives
\begin{equation*}
    \E g_{\vu}(\vX_i)^2
    \leq\E\norm{\vpsi(\vX_i)}_2^4\leq C.
\end{equation*}
In addition,
\begin{equation*}
\begin{aligned}
    0&\leq\E(\vu^\top\vpsi(\vX_i))^2-\E g_{\vu}(\vX_i)\\
    &\leq\E\Big[\norm{\vpsi(\vX_i)}_2^2
       \mathbf1_{\{\norm{\vpsi(\vX_i)}_2^2>A\}}\Big]\\
    &\leq\bigl(\E\norm{\vpsi(\vX_i)}_2^4\bigr)^{1/2}
       \P\bigl(\norm{\vpsi(\vX_i)}_2^2>A\bigr)^{1/2}
       \leq C\frac{\delta^2}{N^2}.
\end{aligned}
\end{equation*}
Apply \cref{lem:sm-blocking} with
$q=\min\{N,\lceil C\log(2N/\delta)\rceil\}$ and $m=\lfloor N/q\rfloor$.
Geometric mixing makes the coupling error at most $\delta/4$.
Also notice that $\log(|\mathcal N|/\delta)\leq Cs\log(P/\delta)$. Therefore with probability at least $1-\delta/2$,
\begin{equation*}
    \max_{\vu\in\mathcal N}
       \left|\frac1N\sum_{i=1}^N
          \bigl(g_{\vu}(\vX_i)-\E g_{\vu}(\vX_i)\bigr)\right|
    \leq C\Big(
       \sqrt{\frac{s\log(P/\delta)}m}
       +\frac{As\log(P/\delta)}m\Big).
\end{equation*}
We conclude the proof by noticing $m^{-1}\leq C\log(2N/\delta)/N$, and applying \cref{lem:sm-restricted-net}.
\end{proof}

The Lasso oracle inequality also requires a bound on the noise score to
set the penalty in \cref{lem:sm-prompt-event-main}. The following lemma uses
a standard exponential supermartingale argument
\citep[Section~4.1, Eq.~(4.3)]{howard2020time} to control this score in
terms of the empirical Gram diagonal.

\begin{lemma}[Martingale noise score]
\label{lem:sm-noise-score}
Fix a task $w$ and suppose \cref{eq:sm-noise-mgf} holds. For any $D>0$ and $0<\delta_0<1$,
\begin{equation}\label{eq:sm-noise-score}
    \P_w\Big(
       \norm{\frac1N\sum_{i=1}^N\vpsi(\vX_i)\xi_i}_\infty
       >\sigma\sqrt{\frac{2D\log(2P/\delta_0)}N},
       \ \max_j(\mG_N)_{jj}\leq D
    \Big)\leq\delta_0.
\end{equation}
\end{lemma}

\begin{proof}[Proof of \cref{lem:sm-noise-score}]
Fix $j\leq P$ and $t\in\R$. Define
$M_k:=\exp\Big(t\sum_{i=1}^k\psi_j(\vX_i)\xi_i-\frac{\sigma^2t^2}{2} \sum_{i=1}^k\psi_j(\vX_i)^2\Big)$.
Let $\mathcal G_k:=\sigma(\vX_1,\ldots,\vX_k,\xi_1,\ldots,\xi_{k-1})$.
Since $M_{k-1}$ and $\psi_j(\vX_k)$ are $\mathcal G_k$-measurable,
\cref{eq:sm-noise-mgf} gives
\begin{equation*}
    \E_w[M_k\mid\mathcal G_k]
    =M_{k-1}\E_w\Big[
       e^{t\psi_j(\vX_k)\xi_k-\frac{\sigma^2t^2}{2}\psi_j(\vX_k)^2}
       \mid\mathcal G_k\Big]
    \leq M_{k-1}.
\end{equation*}
Taking expectations and applying the tower property yields
$\E_w M_k\leq\E_w M_{k-1}\leq\E_w M_0=1$.
For $u,t>0$ and $\sigma>0$, since
$(\mG_N)_{jj}=\frac{1}{N}\sum_{i=1}^N\psi_j(\vX_i)^2$, Markov's inequality give
\begin{equation*}
\begin{split}
    &\P_w\Big(\frac1N\sum_{i=1}^N\psi_j(\vX_i)\xi_i>u,
       \ (\mG_N)_{jj}\leq D\Big)\\
    &\qquad\leq\P_w\Big(M_N>e^{Ntu-\sigma^2t^2ND/2}\Big)
       \leq e^{-Ntu+\sigma^2t^2ND/2}\E_w M_N
       \leq e^{-Ntu+\sigma^2t^2ND/2}.
\end{split}
\end{equation*}
Choosing $t=u/(\sigma^2D)$ gives the upper bound
$e^{-Nu^2/(2\sigma^2D)}$. Applying the same argument to the negative
of each coordinate and taking a union bound over $j=1,\ldots,P$ yields
\begin{equation*}
    \P_w\Big(
       \norm{\frac1N\sum_{i=1}^N\vpsi(\vX_i)\xi_i}_\infty>u,
       \ \max_j(\mG_N)_{jj}\leq D
    \Big)\leq2P e^{-Nu^2/(2\sigma^2D)}.
\end{equation*}
We complete the proof by taking $u=\sigma\sqrt{2D\log(2P/\delta_0)/N}$.
\end{proof}

\subsection{Lasso for Path Dependent Data}


\begin{proof}[Proof of \cref{lem:sm-prompt-event-main}]
Fix a task $w$.
We first bound the diagonal entries of $\mG_N$.
On the event in \cref{eq:sm-prompt-event}, the concentration assumption states that
\begin{equation*}
    |\vv^\top(\mG_N-\E_w[\mG_N])\vv|\leq\alpha,
    \qquad \vv\in\mathcal C(S),\quad\norm{\vv}_2=1.
\end{equation*}
Choosing $k\in S$ and $\vv=\ve_k$ implies $|(\mG_N-\E_w[\mG_N])_{kk}|\leq\alpha$.
For any $j\ne k$, the two unit vectors
$(\ve_k+\ve_j)/\sqrt2$ and $(\ve_k-\ve_j)/\sqrt2$ also belong to
$\mathcal C(S)$. Thus expanding the quadratic forms gives
\begin{equation*}
    \left|\frac{(\mG_N-\E_w[\mG_N])_{kk}+(\mG_N-\E_w[\mG_N])_{jj}}2
       \pm(\mG_N-\E_w[\mG_N])_{kj}\right|\leq\alpha.
\end{equation*}
Therefore,
\begin{equation*}
\begin{aligned}
    |(\mG_N-\E_w[\mG_N])_{kk}+(\mG_N-\E_w[\mG_N])_{jj}|\leq2\alpha,\\
    |(\mG_N-\E_w[\mG_N])_{jj}|\leq2\alpha+|(\mG_N-\E_w[\mG_N])_{kk}|\leq3\alpha.
\end{aligned}
\end{equation*}
Since
$(\E_w[\mG_N])_{jj}\leq B^2$ by \cref{ass:sm-bounded},
we have $\max_j(\mG_N)_{jj}\leq B^2+3\alpha$.
Since $Y_i=f_w(\vX_i)+\xi_i$, we have, for any coordinate $j$,
\begin{equation}\label{eq:sm-decomposition}
\begin{aligned}
\bigl[N^{-1}\mPsi^\top(\vy-\mPsi\vbeta_w^\star)\bigr]_j
=\frac1N\sum_{i=1}^N\psi_j(\vX_i)
   \bigl(f_w(\vX_i)-\vpsi(\vX_i)^\top\vbeta_w^\star\bigr)+\frac1N\sum_{i=1}^N\psi_j(\vX_i)\xi_i.
\end{aligned}
\end{equation}
Applying Cauchy--Schwarz and \cref{ass:sm-sparse} give
\begin{equation}\label{eq:sm-approximation-term}
\begin{aligned}
\Big|\frac1N\sum_{i=1}^N\psi_j(\vX_i)
\bigl(f_w(\vX_i)-\vpsi(\vX_i)^\top\vbeta_w^\star\bigr)\Big|
&\leq
\Big(\frac1N\sum_{i=1}^N\psi_j(\vX_i)^2\Big)^{1/2}
\Big(\frac1N\sum_{i=1}^N
\bigl|f_w(\vX_i)-\vpsi(\vX_i)^\top\vbeta_w^\star\bigr|^2\Big)^{1/2}\\
&\quad\leq\epsilon\sqrt{(\mG_N)_{jj}}
\leq\epsilon\sqrt{B^2+3\alpha}.
\end{aligned}
\end{equation}
The diagonal estimate above allows us to apply \cref{lem:sm-noise-score}
with $D=B^2+3\alpha$ and $\delta_0=(PN)^{-4}$.
Combining its probability bound with \cref{eq:sm-prompt-event} and
absorbing absolute constants yields
\begin{equation}\label{eq:sm-noise-probability}
\begin{aligned}
    \P_w\Big(&\sup_{\vv\in\mathcal C(S),\norm{\vv}_2\leq1}
       |\vv^\top(\mG_N-\E_w[\mG_N])\vv|\leq\alpha,\\
    &\qquad \norm{\frac1N\sum_{i=1}^N\vpsi(\vX_i)\xi_i}_\infty
       \leq C\sigma\sqrt{B^2+3\alpha}\sqrt{\frac{\log(PN)}N}
       \Big)\geq1-C(PN)^{-4}.
\end{aligned}
\end{equation}
This event, together with \cref{eq:sm-decomposition,eq:sm-approximation-term} indicates
\begin{equation}\label{eq:sm-score-bound}
    \norm{N^{-1}\mPsi^\top(\vy-\mPsi\vbeta_w^\star)}_\infty
    \leq\sqrt{B^2+3\alpha}\Big(\epsilon+C\sigma\sqrt{\frac{\log(PN)}{N}}
    \Big)\leq\frac\lambda2.
\end{equation}
By \cref{ass:sm-rsc}, the matrix deviation bound
also implies the empirical RSC
\begin{equation}\label{eq:sm-empirical-rsc}
    \vu^\top\mG_N\vu
    \geq\vu^\top\E_w[\mG_N]\vu-\alpha\norm{\vu}_2^2
    \geq\alpha\norm{\vu}_2^2,
    \qquad \vu\in\mathcal C(S).
\end{equation}
Thus \cref{eq:sm-score-bound,eq:sm-empirical-rsc} both hold on the event in
\cref{eq:sm-noise-probability}.

Let $\vDelta=\vv-\vbeta_w^\star$. The inequality
$J_\lambda(\vv)\leq J_\lambda(\vbeta_w^\star)+\gamma$ implies
\begin{equation*}
\begin{aligned}
\frac12\vDelta^\top\mG_N\vDelta
\leq\left\langle N^{-1}\mPsi^\top(\vy-\mPsi\vbeta_w^\star),\vDelta\right\rangle
+\lambda\bigl(\norm{\vbeta_w^\star}_1
   -\norm{\vbeta_w^\star+\vDelta}_1\bigr)+\gamma.
\end{aligned}
\end{equation*}
By \cref{eq:sm-score-bound} the inner product is bounded by
\begin{equation*}
    \left|\left\langle N^{-1}\mPsi^\top(\vy-\mPsi\vbeta_w^\star),\vDelta\right\rangle\right|
    \leq\norm{N^{-1}\mPsi^\top(\vy-\mPsi\vbeta_w^\star)}_\infty\norm{\vDelta}_1
    \leq\frac\lambda2\norm{\vDelta}_1.
\end{equation*}
Since $(\vbeta_w^\star)_{S^c}=0$, we have $\norm{\vbeta_w^\star}_1-\norm{\vbeta_w^\star+\vDelta}_1\leq \norm{\vDelta_S}_1-\norm{\vDelta_{S^c}}_1$.
Therefore,
\begin{equation*}
    \frac12\vDelta^\top\mG_N\vDelta
    \leq\frac{3\lambda}{2}\norm{\vDelta_S}_1
       -\frac\lambda2\norm{\vDelta_{S^c}}_1+\gamma,
\end{equation*}
which is equivalanet to the Lasso basic inequality
\citep{bickel2009simultaneous,negahban2012unified}:
\begin{equation}\label{eq:sm-basic-inequality}
    \vDelta^\top \mG_N\vDelta+\lambda\norm{\vDelta_{S^c}}_1
    \leq3\lambda\norm{\vDelta_S}_1+2\gamma.
\end{equation}
If $2\gamma\leq\lambda\norm{\vDelta_S}_1$, then $\vDelta\in\mathcal C(S)$ and by RSC,
\begin{equation*}
    \alpha\norm{\vDelta}_2^2
    \leq4\lambda\norm{\vDelta_S}_1
    \leq4\lambda\sqrt{s}\norm{\vDelta}_2.
\end{equation*}
Otherwise, $\lambda\norm{\vDelta_S}_1<2\gamma$, and
\cref{eq:sm-basic-inequality} forces
$\lambda\norm{\vDelta_{S^c}}_1<8\gamma$ and hence
$\norm{\vDelta}_2\leq\norm{\vDelta}_1<10\gamma/\lambda$.
Combining both cases proves \cref{eq:sm-prompt-oracle}.
\end{proof}

\subsection{Verifying Bounded Moments and Concentration Assumptions}\label{app:verify_moments}

We now verify \cref{ass:sm-bounded,ass:sm-concentration} for the stochastic
processes introduced in \cref{sec:sm-examples}. For dissipative diffusions
and the CIR process, we establish signature moment bounds and geometric
mixing to invoke \cref{lem:sm-markov-gram}. For the fractional OU process,
we instead exploit Gaussian concentration and control dependence across
path segments via covariance decay.

Throughout this subsection, 
we fix a task $w$, and all probabilistic quantities and $L^q$ norms are taken under $\P_w$.
\subsubsection{Diffusion with geometric mixing}
\label{app:proof_brownian_sde}

\begin{proof}[Proof of \cref{prop:sm-brownian-sde}]
Since $\vb$ and $\msigma$ are globally Lipschitz,
\cref{eq:sm-brownian-sde} has a unique strong solution.
\paragraph{Signature moments.}
For $g\in\mathcal{C}^2(\R^d)$, the diffusion generator is
$\mathcal Lg=\langle\vb,\nabla g\rangle+\frac12\operatorname{tr}(\msigma\msigma^\top\nabla^2g)$.
We fix $0<a<m/\bar c$ and set $V(\vz)=e^{a\norm{\vz}_2^2}$.
By \cref{eq:sm-sde-coefficients},
\begin{equation*}
    \frac{\mathcal LV(\vz)}{V(\vz)}
    \leq a(2\gamma+d\bar c)-2a(m-a\bar c)\norm{\vz}_2^2.
\end{equation*}
Thus $\mathcal LV\leq-cV+C$ for some constants $c,C>0$.
Applying It\^o's formula to $e^{ct}V(\vZ_t)$, we obtain
\begin{equation}\label{eq:sm-diffusion-exponential-moment}
    \E_w V(\vZ_t)
    \leq e^{-ct}V(\vz_0)+C\int_0^t e^{-c(t-s)}\,\dif s
    \leq V(\vz_0)+C/c.
\end{equation}
Since $\vb$ has linear growth, applying Gr\"onwall's inequality  to the
integral form of \cref{eq:sm-brownian-sde} yields
\begin{equation}\label{eq:sm-bound_z_1}
    \sup_{0\leq t\leq1}\norm{\vZ_{i-1+t}}_2
    \leq C\Big(1+\norm{\vZ_{i-1}}_2+
       \sup_{0\leq t\leq1}
       \norm{\int_{i-1}^{i-1+t}\msigma(\vZ_u)\,\dif\vW_u}_2\Big).
\end{equation}
To control the stochastic integral in \cref{eq:sm-bound_z_1}, we use the Burkholder--Davis--Gundy inequality: for any continuous local martingale $\mathcal M$ starting at zero,
\begin{equation}\label{eq:sm-BDG}
    \norm{\sup_{0\leq t\leq1}|\mathcal M_t|}_{L^q}
    \leq C\sqrt q\,\norm{[\mathcal M]_1^{1/2}}_{L^q},
    \qquad q\geq2,
\end{equation}
with $[\mathcal M]$  its quadratic variation \citep{carlen1991lp}.
Here $\norm{\cdot}_{L^q}$ denotes the $L^q(\P_w)$ norm.
Each coordinate of the stochastic integral in \cref{eq:sm-bound_z_1} has quadratic variation
at most $\bar c$ on a unit interval, so its supremum has $L^q$ norm at most $C\sqrt q$. Likewise, $\norm{\norm{\vZ_{i-1}}_2}_{L^q}\leq C\sqrt q$ uniformly in
$i$, by the exponential moment bound
\cref{eq:sm-diffusion-exponential-moment}.
Inserting these bounds into \cref{eq:sm-bound_z_1}, we get
\begin{equation}\label{eq:sm-diffusion-path-moment}
    \sup_i\norm{\sup_{0\leq t\leq1}\norm{\vZ_{i-1+t}}_2}_{L^q}
    \leq C\sqrt q,\qquad q\geq2.
\end{equation}


{For indices $a_j\in\{0,\ldots,d\}$, let
$S_i^{a_1\cdots a_k}(t)$ denote the corresponding signature coordinate
of $\widetilde{\vX}_i(u)=(u,\vZ_{i-1+u})$ over $[0,t]$, where the index $0$ denotes the augmented time coordinate, and set
$S_i^{\varnothing}(t)=1$. Following the recursive argument in
\citet[Lemma~E.3]{guo2025consistency} and
\citet[Lemma~A.3]{horvath2026fast}, we prove by induction on $k$ that
\begin{equation*}
    \sup_i\norm{\sup_{0\leq t\leq1}|S_i^{a_1\cdots a_k}(t)|}_{L^q}
    \leq C_kq^{k/2},\qquad q\geq2,
\end{equation*}
with $C_k$ independent of $i$ and $q$.
The cases $k=0,1$ follow from $S_i^{\varnothing}(t)=1$,
$S_i^0(t)=t$, and \cref{eq:sm-diffusion-path-moment}.
For $k\geq2$, assume the bound holds at lower levels. If $a_k=0$, the last integration is in time, and the bound improves
to $C_kq^{(k-1)/2}$.
For $a_k\geq1$, write $b_{a_k}$ for the corresponding drift coordinate
and $\msigma_{a_k,\cdot}$ for the corresponding diffusion row.
In It\^o form, the Stratonovich iterated integral reads
\begin{equation}\label{eq:sm-diffusion-signature-ito}
\begin{split}
    S_i^{a_1\cdots a_k}(t)
    &=\int_0^t S_i^{a_1\cdots a_{k-1}}(u)
       b_{a_k}(\vZ_{i-1+u})\,\dif u\\
    &\quad+\int_0^t S_i^{a_1\cdots a_{k-1}}(u)
       \msigma_{a_k,\cdot}(\vZ_{i-1+u})\,\dif\vW_{i-1+u}\\
    &\quad+\frac12\int_0^t S_i^{a_1\cdots a_{k-2}}(u)\,
       \dif[\widetilde X_i^{a_{k-1}},\widetilde X_i^{a_k}]_u.
\end{split}
\end{equation}
Here $[\cdot,\cdot]$ denotes quadratic covariation.
We bound the $L^q$ norm of the supremum over $0\leq t\leq1$
of each term on the right-hand side.
For the drift term, H\"older's inequality and the linear growth of $\vb$ bound the $L^q$
norm by
\begin{equation*}
    C\norm{\sup_{0\leq u\leq1}|S_i^{a_1\cdots a_{k-1}}(u)|}_{L^{2q}}
       \Big(1+\norm{\sup_{0\leq u\leq1}\norm{\vZ_{i-1+u}}_2}_{L^{2q}}\Big)
       \leq C_kq^{(k-1)/2}\cdot\sqrt{q}=C_kq^{k/2}.
\end{equation*}
The martingale term is handled by the BDG inequality \cref{eq:sm-BDG},
which bounds its $L^q$ norm by
\begin{equation*}
    C\sqrt q\,\norm{\Big(\int_0^1
       |S_i^{a_1\cdots a_{k-1}}(u)|^2\,\dif u\Big)^{1/2}}_{L^q}
       \leq C_kq^{k/2}.
\end{equation*}
For the last term, note that it vanishes if $a_{k-1}=0$; otherwise,
$\dif[\widetilde X_i^{a_{k-1}},\widetilde X_i^{a_k}]_u
=(\msigma\msigma^\top)_{a_{k-1},a_k}(\vZ_{i-1+u})\,\dif u$, with coefficient bounded by $\bar c$. Hence its $L^q$ norm is bounded by
\begin{equation*}
    C\norm{\int_0^1|S_i^{a_1\cdots a_{k-2}}(u)|\,\dif u}_{L^q}
       \leq C_kq^{(k-2)/2}.
\end{equation*}
Summing the three bounds completes the
induction.

Each coordinate of $\vpsi(\vX_i)$ is one of
$S_i^{a_1\cdots a_k}(1)$ for $0\leq k\leq M$. Therefore,
\begin{equation}\label{eq:sm-diffusion-signature-moments}
    \sup_i\Big(\E_w\norm{\vpsi(\vX_i)}_2^q\Big)^{1/q}
    \leq C_Mq^{M/2},\qquad q\geq2.
\end{equation}
Taking $q=4$ proves the fourth-moment bound.}


\paragraph{Gram concentration.}
{\hypersetup{linkcolor=orange,citecolor=orange}Let $P_t(\vz,\cdot)$ denote the transition kernel
of the diffusion process $\vZ$ in \cref{eq:sm-brownian-sde}. By the Lipschitz assumptions and
the uniform ellipticity bounds in \cref{eq:sm-sde-coefficients}, $P_t(\vz,\cdot)$ admits a positive, continuous density}
$p_t(\vz,\vz')$ for $t>0$ \citep{menozzi2021density}.
Define $K=\{\vz:V(\vz)\leq r\}$. Since $K$ is compact,
we have $c_K:=\min_{\vz,\vz'\in K}p_1(\vz,\vz')>0$.
The minorization bound below follows from the definition of $c_K$,
and the drift bound from \cref{eq:sm-diffusion-exponential-moment} with
initial state $\vz$:
\begin{equation*}
\begin{aligned}
    P_1(\vz,A)&\geq c_K\operatorname{Leb}(A\cap K),
        &&\vz\in K,\\
    \int V(\vz')\,P_1(\vz,\dif\vz')
        &\leq e^{-c}V(\vz)+C,&&\vz\in\R^d,
\end{aligned}
\end{equation*}
for any Borel set $A$, where $\operatorname{Leb}$ denotes Lebesgue
measure. For $r>\frac{2C}{1-e^{-c}}$, it follows from the geometric Harris
theorem \citep{mattingly2002ergodicity} that there is a unique invariant
probability measure $\pi$, and that
$\norm{P_t(\vz,\cdot)-\pi}_{\TV}\leq CV(\vz)e^{-ct}$ for $t\geq0$.

With these moment and transition bounds, \cref{lem:sm-markov-gram}
applies with $\rho=M/2$ and proves \cref{eq:sm-diffusion-rates}.
Setting $\delta=(PN)^{-4}$, we obtain \cref{eq:sm-prompt-event} whenever
$N\geq Cs[\log P]^{M+2}$.
\end{proof}

\begin{proof}[Proof of \cref{prop:sm-cir-concentration}]
The CIR process has a unique strong
solution \citep{jin2016exponential}.
\paragraph{Signature moments.}
For $g\in C^2([0,\infty))$, the generator is
$\mathcal Lg(z)=\kappa(\mu-z)g'(z)+\frac12\nu^2zg''(z)$.
For $V(z)=e^{az}$ with $0<a<2\kappa/\nu^2$,
\begin{equation*}
    \frac{\mathcal LV(z)}{V(z)}
    =a\kappa\mu-a(\kappa-a\nu^2/2)z,
\end{equation*}
so $\mathcal LV\leq-cV+C$ for some constants $c,C>0$. As in the diffusion case, It\^o's formula shows that
$\sup_t\E_w e^{aZ_t}<\infty$, hence
$\sup_t\norm{Z_t}_{L^q}\leq Cq$ for $q\geq1$.
Similar to \cref{eq:sm-bound_z_1} and by Burkholder--Davis--Gundy inequality, we obtain
\begin{equation*}
    \norm{\sup_{0\leq t\leq1}Z_{i-1+t}}_{L^q}
    \leq Cq+C\sqrt q\Big(\int_{i-1}^i\norm{Z_u}_{L^{q/2}}\,\dif u\Big)^{1/2}
    \leq Cq,\qquad q\geq2.
\end{equation*}

For a one-dimensional path $x$, the Stratonovich signature has a simple bound
$\norm{\vpsi(x)}_2\leq C_M(1+\norm{x}_\infty)^M$. Consequently,
\begin{equation}\label{eq:sm-cir-signature-moments}
\sup_i\Big(\E_w\norm{\vpsi(\vX_i)}_2^q\Big)^{1/q}
\leq C_M\Big(1+\sup_i\norm{\sup_{0\leq t\leq1}Z_{i-1+t}}_{L^{Mq}}\Big)^M
\leq C_Mq^M,\quad q\geq2.
\end{equation}
Taking $q=4$ proves the fourth-moment bound.
\paragraph{Gram concentration.}
The CIR transition law satisfies
$\norm{P_t(z,\cdot)-\pi}_{\TV}
\leq C(1+z)e^{-ct}$, where $\pi$ is the gamma distribution \citep{jin2016exponential}.
Since $\sup_t\E_w Z_t<\infty$, \cref{lem:sm-markov-gram} applies with
$V(z)=1+z$ and $\rho=M$, which proves \cref{eq:sm-cir-rates}.
With $\delta=(PN)^{-4}$, \cref{eq:sm-prompt-event} follows for
$N\geq Cs[\log P]^{2M+2}$.
\end{proof}

\subsubsection{Fractional OU process}
\label{app:proof_fou_concentration}

\begin{proof}[Proof of \cref{prop:sm-fou-concentration}]
Let $\overline{Z}$ be the stationary fractional OU process
driven by a two-sided extension of the same fractional Brownian motion.
By \citet[Section~2]{cheridito2003fractional},
\begin{equation}\label{eq:sm-fou-stationary}
    Z_t=\overline{Z}_t+e^{-\kappa t}(z_0-\overline{Z}_0).
\end{equation}
\paragraph{Signature moments.}
By Gaussian concentration for the supremum of $\overline{Z}$
\citep{ledoux2001concentration} and
{\cref{eq:sm-fou-stationary}},
we have $\sup_i\P_w(\norm{\vX_i}_\infty>u)\leq Ce^{-cu^2}$ for $u\geq C$.
{ Indeed,
$\norm{\vX_i}_\infty\leq\norm{\overline Z_{i-1+\cdot}}_\infty
+|z_0|+|\overline Z_0|$, and stationarity yields a uniform  tail bound uniform in $i$.}
For a scalar path, recall that signature coordinates can be expressed using products of increments and time integrals, hence
$\norm{\vpsi(x)}_2\leq C_M(1+\norm{x}_\infty)^M$, and the fourth-moment bound
follows from the Gaussian tail bound. 
{The same representation also gives a local Lipschitz bound:
for $\norm{x}_\infty,\norm{y}_\infty\leq A$,}
\begin{equation}\label{eq:sm-fou-lipschitz}
    \norm{\vpsi(x)-\vpsi(y)}_2
    \leq C_M(1+A)^{M-1}\norm{x-y}_\infty.
\end{equation}

\paragraph{Gram concentration.}
By \citet[Theorem~2.3]{cheridito2003fractional}, the stationary
covariance $R_H(h):=\operatorname{Cov}(\overline{Z}_0,\overline{Z}_h)$,
$h\geq0$, satisfies
\begin{equation}\label{eq:sm-fou-covariance-decay}
    |R_H(h)|\leq C(1+h)^{2H-2}.
\end{equation}
{And for $H=1/2$, the covariance can be explicitly computed:
$R_{1/2}(h)=\frac{\nu^2}{2\kappa}e^{-\kappa h}$.}
We will verify the condition by applying Gaussian concentration to the $N$ sampled unit-time paths
$\vX_i(t)=Z_{i-1+t}$, viewed jointly as a Gaussian process
$(\vX_1,\ldots,\vX_N)\in C([0,1])^N$.

We first bound the variances of linear functionals on this product space. 
Let
$a_{ij}:=\sup_{s,t\in[0,1]}
\left|\operatorname{Cov}(\overline{Z}_{i-1+s},\overline{Z}_{j-1+t})\right|$. Then we have
{
\begin{equation*}
    \max_i\sum_{j=1}^N a_{ij}\leq CN^{\max(2H,1)-1}.
\end{equation*}}
Consider the path space $C([0,1])^N$ equipped with the norm
$\norm{\vx}:=(\sum_{i=1}^N\norm{x_i}_\infty^2)^{1/2}$ for
$\vx=(x_1,\ldots,x_N)$. By the Riesz representation theorem, any
continuous linear functional in this space can be written in the form of $L_\mu(\vx)=\sum_{i=1}^N\int_0^1x_i(t)\,\dif\mu_i(t)$ for
some finite signed measures $\mu_i$ on $[0,1]$. Writing
$m_i:=|\mu_i|([0,1])$ for their total variations, we have
$\norm{L_\mu}_{\op}^2=\sum_{i=1}^N m_i^2$, and
\begin{equation}\label{eq:sm-var_fou_stationary}
\begin{aligned}
\operatorname{Var}\Big(\sum_{i=1}^N\int_0^1
\overline{Z}_{i-1+t}\,\dif\mu_i(t)\Big)
&=\sum_{i,j=1}^N\int_0^1\int_0^1
\operatorname{Cov}(\overline{Z}_{i-1+s},\overline{Z}_{j-1+t})
\,\dif\mu_i(s)\,\dif\mu_j(t)\\
&\leq\sum_{i=1}^N m_i^2\sum_{j=1}^N a_{ij}\\
&\leq CN^{\max(2H,1)-1}\sum_{i=1}^N m_i^2.
\end{aligned}
\end{equation}
By \cref{eq:sm-fou-stationary}, $L_\mu(\vX_1,\ldots,\vX_N)=L_\mu(\overline{\vX}_1,\ldots,\overline{\vX}_N)+c_\mu(z_0-\overline{Z}_0)$, where
$c_\mu:=\sum_{i=1}^N\int_0^1e^{-\kappa(i-1+t)}\,\dif\mu_i(t)$ is deterministic.
Since $z_0$ is fixed, {the correction term $c_\mu(z_0-\overline Z_0)$} has variance
$c_\mu^2\operatorname{Var}(\overline{Z}_0)$, and
\begin{equation}\label{eq:sm-bound_c_mu}
|c_\mu|^2
\leq\big(\sum_{i=1}^N e^{-\kappa(i-1)}m_i\big)^2
\leq\big(\sum_{i=1}^N e^{-2\kappa(i-1)}\big)\sum_{i=1}^N m_i^2
 \leq C\sum_{i=1}^N m_i^2.
\end{equation}
Putting \cref{eq:sm-var_fou_stationary,eq:sm-bound_c_mu} together, we
get
\begin{equation}\label{eq:sm-fou-covariance-operator}
\begin{split}
    \operatorname{Var}\Big(L_\mu(\vX_1,\ldots,\vX_N)\Big)
    &\leq2\operatorname{Var}\Big(\sum_{i=1}^N\int_0^1
       \overline{Z}_{i-1+t}\,\dif\mu_i(t)\Big)
       +2c_\mu^2\operatorname{Var}(\overline{Z}_0)\\
    &\leq CN^{\max(2H,1)-1}\sum_{i=1}^N m_i^2\\
    &=CN^{\max(2H,1)-1}\norm{L_\mu}_{\op}^2.
\end{split}
\end{equation}
Let $\mathcal N$ be the finite set of unit vectors from
\cref{lem:sm-restricted-net}, which reduces the supremum on the Lasso cone to a maximum over $\mathcal N$.
For $\vu\in\mathcal N$, define
$F_{\vu}(\vx)=\frac1N\sum_{i=1}^N(\vu^\top\vpsi(x_i))^2$. Then
\begin{equation*}
    F_{\vu}(\vX_1,\ldots,\vX_N)
    -\E_w F_{\vu}(\vX_1,\ldots,\vX_N)
    =\vu^\top(\mG_N-\E_w[\mG_N])\vu.
\end{equation*}

{Combining \cref{eq:sm-fou-lipschitz} with
$\norm{\vpsi(x)}_2\leq C_M(1+\norm{x}_\infty)^M$, we obtain
\begin{equation*}
\begin{aligned}
    |F_{\vu}(\vx)-F_{\vu}(\vy)|
    &\leq\frac{C_M(1+A)^{2M-1}}{N}
       \sum_{i=1}^N\norm{x_i-y_i}_\infty\\
    &\leq\frac{C_M(1+A)^{2M-1}}{\sqrt N}\norm{\vx-\vy}.
\end{aligned}
\end{equation*}
Using $\sup_i\P_w(\norm{\vX_i}_\infty>u)\leq Ce^{-cu^2}$,
a union bound allows us to choose
$A=C\sqrt{\log(2N/\delta)}$ such that
\begin{equation*}
    \P_w\bigl(\max_i\norm{\vX_i}_\infty>A\bigr)
    \leq CN e^{-cA^2}\leq\frac{\delta^2}{4N^2}.
\end{equation*}
Clipping each path to $[-A,A]$ extends the Lipschitz bound to the whole
path space and leaves $F_{\vu}(\vx)$ unchanged whenever
$\max_i\norm{x_i}_\infty\leq A$.}

Define the coordinate-wise clipping $\vx^A(t):=\max\{-A,\min\{\vx(t),A\}\}$, and $\widetilde F_{\vu}(\vx):=F_{\vu}(\vx^A)$.
Consequently, $\widetilde F_{\vu}$ is
$C_M(1+A)^{2M-1}/\sqrt N$-Lipschitz on $C([0,1])^N$ and satisfies
$0\leq\widetilde F_{\vu}(\vx)\leq C_M(1+A)^{2M}$.
By the Gaussian concentration \citep{ledoux2001concentration} and the
variance bound \cref{eq:sm-fou-covariance-operator},  we see
\begin{equation}\label{eq:sm-fou-gaussian-concentration}
    \P_w\bigl(|\widetilde F_{\vu}-\E_w\widetilde F_{\vu}|>t\bigr)
    \leq2\exp\Big(-\frac{cN^{2-\max(2H,1)}t^2}{(1+A)^{4M-2}}\Big).
\end{equation}
Since $\widetilde F_{\vu}=F_{\vu}$ on the event
$\{\max_i\norm{\vX_i}_\infty\leq A\}$ and $\E_w F_{\vu}^2\leq B^4$,
we see by Cauchy--Schwarz inequality that the truncation bias satisfies
$|\E_w F_{\vu}-\E_w\widetilde F_{\vu}|\leq
C\bigl(B^2\delta/N+(1+A)^{2M}\delta^2/N^2\bigr)$. 
Now a union bound over $\vu\in\mathcal N$ in
\cref{eq:sm-fou-gaussian-concentration}, followed by
\cref{lem:sm-restricted-net}, yileds \cref{eq:sm-fou-rates}. Finally, with $\delta=(PN)^{-4}$ and a sufficiently large constant in \cref{eq:sm-fou-sample-size}, we obtain \cref{eq:sm-prompt-event}.
\end{proof}

\section{Construction of the Transformer}
\label{sec:transfer-construction}

This section constructs the masked softmax transformer of \cref{lem:sm-transformer-lasso-main} by assembling three modular building blocks. 
First, \cref{lem:sm-softmax-gradient} shows that two masked softmax attention heads can approximate a gradient step on the empirical
squared loss. 
Second, \cref{lem:sm-threshold} constructs an MLP block that executes the proximal update via soft-thresholding and clipping. 
Third, \cref{lem:sm-readout} designs a dedicated readout attention layer that extracts the scalar query prediction from the final coefficient iterate. 
Finally, these components are stacked across layers to solve in-context Lasso up to any prescribed
optimization tolerance.

\subsection{Masked Softmax Attention Realization of Proximal Gradient}

We now present the first and the core transformer construction: a masked softmax attention that approximates in-context proximal gradient.

To start, let $\vX_1,\ldots,\vX_{N+1}\in\mathcal X$ be paths with observed labels
$Y_1,\ldots,Y_N\in\R$ for the first $N$ paths. Their signature features are
$\vpsi(\vX_i)=\overline{\Sig}_M(\widetilde{\vX}_i)\in\R^P$.
For $\lambda>0$, define the Lasso objective
\begin{equation}\label{eq:sm-deterministic-objective}
    J_\lambda(\vu):=\frac1{2N}\sum_{i=1}^N(\vpsi(\vX_i)^\top\vu-Y_i)^2
       +\lambda\norm{\vu}_1.
\end{equation}
We write a token embedding as $\vh=[\vx;y;\vv;b_{\rm d};b_{\rm q}]$, where
$\vx\in\R^P$ is its feature vector and $b_{\rm d}$ and $b_{\rm q}$
indicate demonstration and query tokens, respectively.
We encode the path-label pairs and the query, together with a zero anchor, as
\begin{equation}\label{eq:sm-token-encoding}
\begin{aligned}
    \vh_0&=[\vzero_P;0;\vzero_P;0;0],\\
    \vh_i&=[\vpsi(\vX_i);Y_i;\vzero_P;1;0],\qquad 1\leq i\leq N,\\
    \vh_{N+1}&=[\vpsi(\vX_{N+1});0;\vv;0;1].
\end{aligned}
\end{equation}
The coefficient vector $\vv$ is initially zero and stores the
successive coefficient estimates. 

\begin{lemma}[Gradient step by attention]
\label{lem:sm-softmax-gradient}
Fix a truncation radius $K>0$ and let $\eta,R,\gamma>0$. There is a two-head masked softmax attention
block $\Attn_\vtheta(\cdot)$ (\cref{def:masked-softmax-attention}) with head dimension $P+1$ satisfying the following property. Let $\mH=[\vh_0,\ldots,\vh_{N+1}]^\top$ be a prompt of the
form \cref{eq:sm-token-encoding} such that
\begin{equation*}
    \max_{1\leq i\leq N+1}\norm{\vpsi(\vX_i)}_2
       \vee\max_{1\leq i\leq N}|Y_i|\leq K,
    \qquad \norm{\vv}_\infty\leq R.
\end{equation*}
Then $\Attn_\vtheta(\cdot)$ leaves the first $N+1$ tokens unchanged and maps
$\vh_{N+1}$ to $\widetilde\vh_{N+1}=[\vpsi(\vX_{N+1});0;\widetilde\vv;0;1]$,
where $\widetilde\vv$ is within $\gamma$ of one gradient step on the
squared loss:
\begin{equation}\label{eq:sm-one-step-error}
    \norm{\widetilde \vv-
    \big[\vv-\frac\eta N\sum_{i=1}^N(\vv^\top \vpsi(\vX_i)-Y_i)\vpsi(\vX_i)\big]}_2
    \leq\gamma.
\end{equation}
\end{lemma}

\begin{proof}
Let us take a generic token as $\vh=[\vx;y;\vv;b_{\rm d};b_{\rm q}]$ and refer to the two heads as the positive and negative heads, indexing their parameters by $+$ and $-$. The heads share a query, and have opposite keys and
values: for $\tau,A>0$, set
\begin{equation}\label{eq:sm-gradient-heads}
\begin{aligned}
    \mQ_\pm^\top \vh&=[\vv;b_{\rm q}],\qquad
    \mK_\pm^\top \vh=[\pm\tau \vx;\mp\tau y-Ab_{\rm d}-2Ab_{\rm q}],\\
    \mV_\pm^\top \vh&=\mp\frac{\eta e^A}{2\tau N}[\vzero_P;0;\vx;0;0].
\end{aligned}
\end{equation}
Since the values are supported on the coefficient coordinates, every token keeps all
its other coordinates due to residual connection.
At the anchor and demonstration tokens, $\vv=\vzero_P$ and
$b_{\rm q}=0$, so the query vector $[\vv;b_{\rm q}]$ vanishes and thus all attention scores are zero. Consequently, both heads place uniform weights on the visible tokens, and their opposite values cancel. These tokens are therefore unchanged.

For the last token $\vh_{N+1}$, the query vector is $[\vv;1]$. Write
$r_i=\vv^\top\vpsi(\vX_i)-Y_i$ for the residual of demonstration $i$,
and $z=\vv^\top\vpsi(\vX_{N+1})$. By \cref{eq:sm-gradient-heads}, each
head assigns to token $\vh_s$ the score
\begin{equation*}
    \langle[\vv;1],\mK_\pm^\top\vh_s\rangle=
    \begin{cases}
       0, & s=0\ \text{(anchor)},\\
       -A\pm\tau r_s, & 1\leq s\leq N\ \text{(demonstrations)},\\
       -2A\pm\tau z, & s=N+1\ \text{(itself)}.
    \end{cases}
\end{equation*}
By \cref{eq:masked-softmax-attention-vector}, the positive head
therefore adds to the coefficient vector of $\vh_{N+1}$
\begin{equation*}
-\frac{\eta e^A}{2\tau N}\cdot
\frac{\sum_{i=1}^N e^{-A+\tau r_i}\vpsi(\vX_i)
 +e^{-2A+\tau z}\vpsi(\vX_{N+1})}
{1+\sum_{i=1}^N e^{-A+\tau r_i}+e^{-2A+\tau z}}.
\end{equation*}
The negative head contributes the same expression with
$\tau$ replaced by $-\tau$. Adding both contributions to the input
$\vv$, we obtain
\begin{equation}\label{eq:sm-gradient-update}
\begin{aligned}
    \widetilde\vv
    &=\vv-\frac\eta{2\tau N}
       \frac{\sum_{i=1}^N e^{\tau r_i}\vpsi(\vX_i)+e^{-A+\tau z}\vpsi(\vX_{N+1})}
            {1+e^{-A}\sum_{i=1}^N e^{\tau r_i}+e^{-2A+\tau z}}\\
    &\quad+\frac\eta{2\tau N}
       \frac{\sum_{i=1}^N e^{-\tau r_i}\vpsi(\vX_i)+e^{-A-\tau z}\vpsi(\vX_{N+1})}
            {1+e^{-A}\sum_{i=1}^N e^{-\tau r_i}+e^{-2A-\tau z}}.
\end{aligned}
\end{equation}
It remains to show that this update is close to the gradient step with suitable parameters $A,\tau$. Since $\norm{\vpsi(\vX_i)}_2\leq K$, 
$|z|\leq KR\sqrt P$, and $|r_i|\leq K(R\sqrt P+1)$, the denominators
in \cref{eq:sm-gradient-update} are close to one, and the terms
involving $\vpsi(\vX_{N+1})$ are small: for $0<\tau\leq1$ and $A>0$,
\begin{equation*}
\begin{aligned}
    0\leq1-\frac{1}
       {1+e^{-A}\sum_{i=1}^N e^{\pm\tau r_i}+e^{-2A\pm\tau z}}
       &\leq Ce^{-A},\\
    e^{-A\pm\tau z}\norm{\vpsi(\vX_{N+1})}_2
       &\leq Ce^{-A},
\end{aligned}
\end{equation*}
where $C$ denotes a constant depending only on
$N,P,K,R,\eta$. Moreover, for the terms involving $\vpsi(\vX_{i})$, the symmetric difference quotient is close to $r_i$:
\begin{equation*}
    \Big|\frac{e^{\tau r_i}-e^{-\tau r_i}}{2\tau}-r_i\Big|
    \leq\frac{\tau^2|r_i|^3}{6}e^{\tau|r_i|}
    \leq C\tau^2.
\end{equation*}
Inserting these estimates into \cref{eq:sm-gradient-update}, we find
that
\begin{equation*}
    \norm{\widetilde\vv-
       \big[\vv-\frac\eta N\sum_{i=1}^N r_i\vpsi(\vX_i)\big]}_2
    \leq C\Big(\tau^2+\frac{e^{-A}}\tau\Big)
\end{equation*}
uniformly over the prompts and coefficient vectors in the lemma.
By first fixing $\tau\in(0,1]$ with $C\tau^2\leq\gamma/2$, and then taking $A>0$ with $Ce^{-A}/\tau\leq\gamma/2$, we obtain \cref{eq:sm-one-step-error}.

\end{proof}

\begin{lemma}[Soft-thresholding by MLP]
\label{lem:sm-threshold}
Fix $t,R>0$. There is an MLP block $\MLP_\vtheta(\cdot)$
(\cref{def:mlp}) with hidden width $4P+1$ satisfying the following
property. For any prompt $\mH$ of the form \cref{eq:sm-token-encoding},
the block leaves $\vh_0,\ldots,\vh_N$ unchanged and maps the last token $\vh_{N+1}$ to $\widetilde\vh_{N+1}=[\vpsi(\vX_{N+1});0;\widetilde\vv;0;1]$, where
\begin{equation}\label{eq:sm-soft-threshold}
\widetilde\vv=\operatorname{proj}_{[-R,R]^P}\soft_t(\vv),\qquad
[\soft_t(\vv)]_j:=\operatorname{sign}(v_j)(|v_j|-t)_+.
\end{equation}
Moreover, the map $\vv\mapsto\widetilde\vv$ is $1$-Lipschitz in $\norm{\cdot}_2$.
\end{lemma}

\begin{proof}
Write a token as $\vh=[\vx;y;\vv;b_{\rm d};b_{\rm q}]\in\R^{2P+3}$.
Let $\mE\in\R^{(2P+3)\times P}$ be the matrix with $\mE^\top\vh=\vv$,
and let $\ve_{\rm q}\in\R^{2P+3}$
be the standard basis vector with $\ve_{\rm q}^\top\vh=b_{\rm q}$.  We define $\mW_1, \mW_2 \in \R^{(2P+3)\times(4P+1)}$ by
\begin{equation*}
\begin{aligned}
    \mW_1&=\big[\mE+t\ve_{\rm q}\vone_P^\top,\ \mE-t\ve_{\rm q}\vone_P^\top,\
       \ve_{\rm q},\ \mE-(t+R)\ve_{\rm q}\vone_P^\top,\
       -\mE-(t+R)\ve_{\rm q}\vone_P^\top\big],\\
    \mW_2&=\mE\big[-\mI_P,\ \mI_P,\ t\vone_P,\ -\mI_P,\ \mI_P\big],
\end{aligned}
\end{equation*}
Then, for any token $\vh$, we have
\begin{equation*}
    \mW_1^\top\vh=\big[\vv+tb_{\rm q}\vone_P;\ \vv-tb_{\rm q}\vone_P;\
       b_{\rm q};\ \vv-(t+R)b_{\rm q}\vone_P;\
       -\vv-(t+R)b_{\rm q}\vone_P\big].
\end{equation*}
By \cref{def:mlp}, the MLP block maps $\vh$ to
$\vh+\mW_2\operatorname{ReLU}(\mW_1^\top\vh)$. Since the columns of $\mW_2$ lie in the column space of $\mE$, only the coefficient coordinates are updated, with $\vv$ mapped to
\begin{equation}\label{eq:sm-mlp_update_v}
\begin{aligned}
    \widetilde{\vv}&=\vv-\operatorname{ReLU}(\vv+tb_{\rm q}\vone_P)
     +\operatorname{ReLU}(\vv-tb_{\rm q}\vone_P)
     +t\operatorname{ReLU}(b_{\rm q})\vone_P\\
     &\qquad -\operatorname{ReLU}(\vv-(t+R)b_{\rm q}\vone_P)
     +\operatorname{ReLU}(-\vv-(t+R)b_{\rm q}\vone_P).
\end{aligned}
\end{equation}
For the anchor and demonstration tokens, we have $b_{\rm q}=0$ and $\vv=\vzero_P$; consequently, all ReLU activations vanish and these tokens remain unchanged.

At the last token $\vh_{N+1}$, we have $b_{\rm q}=1$. The first line in \cref{eq:sm-mlp_update_v} then reduces to $\soft_t(\vv)$ via the scalar identity
\begin{equation*}
    a-(a+t)_++(a-t)_++t
       =\operatorname{sign}(a)(|a|-t)_+.
\end{equation*}
The second line subsequently clips this output to $[-R,R]^P$: for each coordinate $j$, its two terms vanish when $\vert{}v_j\vert{}\leq t+R$, and otherwise saturate the soft-thresholded value at $\operatorname{sign}(v_j)R$.
This proves \cref{eq:sm-soft-threshold}. Finally, the map $\vv\mapsto\widetilde\vv$ in \cref{eq:sm-soft-threshold} acts coordinatewise, and each coordinate map $a\mapsto\operatorname{proj}_{[-R,R]}\bigl(\operatorname{sign}(a)(|a|-t)_+\bigr)$ is continuous and piecewise linear with slopes in $[0,1]$, hence $1$-Lipschitz. Therefore the map $\vv\mapsto\widetilde\vv$ is $1$-Lipschitz with respect to $\norm{\cdot}_2$.
\end{proof}

\begin{lemma}[Label readout]
\label{lem:sm-readout}
Fix a truncation radius $K>0$ and let $R,\gamma>0$. There is a two-head masked softmax attention
block $\Attn_\vtheta(\cdot)$ (\cref{def:masked-softmax-attention})
with head dimension $P+1$ satisfying the following property.
Let $\mH$ be a prompt of the form \cref{eq:sm-token-encoding} such that
\begin{equation*}
    \norm{\vpsi(\vX_{N+1})}_2\leq K,
    \qquad \norm{\vv}_\infty\leq R.
\end{equation*}
Then the block leaves $\vh_0,\ldots,\vh_N$ unchanged and maps the last token $\vh_{N+1}$ to $\widetilde\vh_{N+1}=[\vpsi(\vX_{N+1});\widehat y_{\rm q};\vv;0;1]$, where
\begin{equation}\label{eq:sm-readout-error}
    |\widehat y_{\rm q}-\vpsi(\vX_{N+1})^\top\vv|\leq\gamma.
\end{equation}
For any query path $\vX_{N+1}$ and coefficient vector $\vv$, the same
block also satisfies
$|\widehat y_{\rm q}|\leq|\vpsi(\vX_{N+1})^\top\vv|$.
\end{lemma}

\begin{proof}
Write a generic token as $\vh=[\vx;y;\vv;b_{\rm d};b_{\rm q}]$. As in the proof of \cref{lem:sm-softmax-gradient}, we use a positive and a
negative head, and the upper sign in $\pm$ refers to the positive head.
For parameters $\tau,A>0$, unrelated to those in
\cref{lem:sm-softmax-gradient}, set
\begin{equation}\label{eq:sm-readout-heads}
\begin{aligned}
    \mQ_\pm^\top \vh&=[\vx;b_{\rm q}],\qquad
    \mK_\pm^\top \vh=[\pm\tau \vv;-Ab_{\rm d}],\\
    \mV_\pm^\top \vh&=\pm\frac2\tau[\vzero_P;b_{\rm q};\vzero_P;0;0].
\end{aligned}
\end{equation}
The values are supported on the label coordinate and vanish at every
token except $\vh_{N+1}$, the only token with $b_{\rm q}=1$. By the causal mask, each of $\vh_0,\ldots,\vh_N$ attends only to tokens within $\vh_0,\ldots,\vh_N$, all of which have zero value projections, leaving $\vh_0,\ldots,\vh_N$ unchanged.

For the last token $\vh_{N+1}$, the query vector of both heads is
$[\vpsi(\vX_{N+1});1]$. Write $z=\vpsi(\vX_{N+1})^\top\vv$. By
\cref{eq:sm-readout-heads}, each head assigns to token $\vh_s$ the score
\begin{equation*}
    \langle[\vpsi(\vX_{N+1});1],\mK_\pm^\top\vh_s\rangle=
    \begin{cases}
       0, & s=0\ \text{(anchor)},\\
       -A, & 1\leq s\leq N\ \text{(demonstrations)},\\
       \pm\tau z, & s=N+1\ \text{(self)}.
    \end{cases}
\end{equation*}
Only $\vh_{N+1}$ has a nonzero value with $\mV_\pm^\top\vh_{N+1}=\pm\frac2\tau[\vzero_P;1;\vzero_P;0;0]$, and its
label coordinate is initially zero. Summing the outputs of the two heads
in \cref{eq:masked-softmax-attention-vector}, we obtain the new label
\begin{equation*}
    \widehat y_{\rm q}
    =\frac2\tau\Big[
       \frac{e^{\tau z}}{1+Ne^{-A}+e^{\tau z}}
       -\frac{e^{-\tau z}}{1+Ne^{-A}+e^{-\tau z}}
    \Big].
\end{equation*}
Equivalently, $\widehat y_{\rm q}=\frac2\tau[\phi(\tau z)-\phi(-\tau z)]$, where $\phi(a)=e^a/(1+Ne^{-A}+e^a)$. Since $0<\phi'(a)=\phi(a)(1-\phi(a))\leq1/4$ for all $a\in\R$, the mean value theorem implies that
\begin{equation*}
    |\widehat y_{\rm q}|\leq\frac2\tau\cdot\frac14\,|2\tau z|=|z|=|\vpsi(\vX_{N+1})^\top\vv|.
\end{equation*}
To establish \cref{eq:sm-readout-error}, we compare $\phi$ with its limit
$\phi_\infty(a)=e^a/(1+e^a)$ as $A\to\infty$. For any $a\in\R$,
\begin{equation*}
    0\leq\phi_\infty(a)-\phi(a)
    =\frac{Ne^{-A}e^a}{(1+e^a)(1+Ne^{-A}+e^a)}
    \leq Ne^{-A}.
\end{equation*}
Since $\phi_\infty(a)-\phi_\infty(-a)=\tanh(a/2)$, it follows that
\begin{equation*}
    \Big|\widehat y_{\rm q}-\frac2\tau\tanh\Big(\frac{\tau z}2\Big)\Big|
    \leq\frac{4Ne^{-A}}\tau.
\end{equation*}
Moreover, since $|\tanh x-x|\leq|x|^3/3$ for all $x\in\R$, and
$|z|\leq\norm{\vpsi(\vX_{N+1})}_2\norm{\vv}_2\leq KR\sqrt P$ by assumptions, we obtain 
\begin{equation*}
    |\widehat y_{\rm q}-z|
    \leq\frac{4Ne^{-A}}\tau+\frac{\tau^2(KR\sqrt P)^3}{12}.
\end{equation*}
First choosing $\tau>0$ such that $\tau^2(KR\sqrt P)^3\leq6\gamma$, and then choosing $A>0$ such that $4Ne^{-A}\leq\tau\gamma/2$, yields \cref{eq:sm-readout-error}.
\end{proof}

\subsection{Transformers Solve In-Context Lasso}\label{app:proof_transformer}
\begin{proof}[Proof of \cref{lem:sm-transformer-lasso-main}]
Set $\eta=(1+K^2)^{-1}$. The empirical gram matrix
$\mG_N=N^{-1}\sum_i\vpsi(\vX_i)\vpsi(\vX_i)^\top$ satisfies $0\preceq\mG_N\preceq K^2\mI_P$.  Starting from
$\vu^{(0)}=\vzero_P$, let us consider the exact proximal-gradient iterates
\begin{equation*}
    \vu^{(\ell+1)}=\operatorname{proj}_{[-R,R]^P}
       \soft_{\eta\lambda}\Big(
          \vu^{(\ell)}-\frac\eta N\sum_{i=1}^N
             (\vpsi(\vX_i)^\top\vu^{(\ell)}-Y_i)\vpsi(\vX_i)\Big).
\end{equation*}
Then, applying the standard proximal-gradient convergence rate \citep[Theorem~10.21]{beck2017first} yields
\begin{equation}\label{eq:sm-ista-rate}
    J_\lambda(\vu^{(L)})-\min_{\vu\in[-R,R]^P}J_\lambda(\vu)
    \leq\frac{R^2P}{2\eta L}\leq\frac\gamma2,
\end{equation}
where the last inequality follows by taking $L=\lceil R^2P/(\eta\gamma)\rceil$.

Now, define the gradient map and the proximal clipping map
\begin{equation*}
    G(\vu):=\vu-\frac\eta N\sum_{i=1}^N(\vpsi(\vX_i)^\top\vu-Y_i)\vpsi(\vX_i),
    \qquad
    \operatorname{prox}(\vu):=\operatorname{proj}_{[-R,R]^P}\soft_{\eta\lambda}(\vu),
\end{equation*}
so that the proximal-gradient iterates satisfy
$\vu^{(\ell+1)}=\operatorname{prox}(G(\vu^{(\ell)}))$.
Since $0\preceq\mG_N\preceq\eta^{-1}\mI_P$, we have
$\norm{\mI_P-\eta\mG_N}_{\op}\leq1$. This shows that both maps are nonexpansive:
for all $\vu,\vu'\in\R^P$,
\begin{equation*}
    \norm{G(\vu)-G(\vu')}_2
    =\norm{(\mI_P-\eta\mG_N)(\vu-\vu')}_2\leq\norm{\vu-\vu'}_2,
    \quad
    \norm{\operatorname{prox}(\vu)-\operatorname{prox}(\vu')}_2
    \leq\norm{\vu-\vu'}_2.
\end{equation*}
Fix $\gamma_0>0$ (to be specified later). Then each transformer layer composes the attention block of \cref{lem:sm-softmax-gradient} (with one-step error $\gamma_0$) and the MLP block of \cref{lem:sm-threshold} (with threshold $t=\eta\lambda$). For $\ell\geq0$, let $\vv^{(\ell)}$ denote the coefficient vector of $\vh_{N+1}$ after $\ell$ layers, initialized at $\vv^{(0)}=\vzero_P$, and let $\widetilde\vv^{(\ell+1)}$ denote this vector immediately after the attention block of layer $\ell+1$, so that $\vv^{(\ell+1)}=\operatorname{prox}(\widetilde\vv^{(\ell+1)})$. By \cref{lem:sm-threshold}, we have $\norm{\vv^{(\ell)}}_\infty\leq R$ for all $\ell\geq0$. Thus, the conditions in \cref{lem:sm-softmax-gradient} hold at every layer, yielding
\begin{equation*}
\norm{\widetilde\vv^{(\ell+1)}-G(\vv^{(\ell)})}_2\leq\gamma_0
\qquad \text{for all } \ell\geq0.
\end{equation*}

On $[-R,R]^P$, the gradient of the squared-loss term in $J_\lambda$
has Euclidean norm at most
$K^2(R\sqrt P+1)$, and $\norm{\vu-\vv}_1\leq\sqrt P\norm{\vu-\vv}_2$.
Hence
\begin{equation}\label{eq:sm-objective-lipschitz}
    |J_\lambda(\vu)-J_\lambda(\vv)|
    \leq\bigl[K^2(R\sqrt P+1)+\lambda\sqrt P\bigr]\norm{\vu-\vv}_2,
    \qquad \vu,\vv\in[-R,R]^P.
\end{equation}
We take $\gamma_0=\gamma/\{2L[K^2(R\sqrt P+1)+\lambda\sqrt P]\}$.
Since $\vv^{(\ell+1)}-\vu^{(\ell+1)}
=\operatorname{prox}(\widetilde\vv^{(\ell+1)})
-\operatorname{prox}(G(\vu^{(\ell)}))$, the two Lipschitz bounds and the
triangle inequality show that, for $\ell=0,\ldots,L-1$,
\begin{equation}\label{eq:sm-iterate-error}
    \norm{\vv^{(\ell+1)}-\vu^{(\ell+1)}}_2
    \leq\norm{\vv^{(\ell)}-\vu^{(\ell)}}_2
       +\frac{\gamma}{2L\bigl[K^2(R\sqrt P+1)+\lambda\sqrt P\bigr]}.
\end{equation}
Starting from $\vv^{(0)}=\vu^{(0)}=\vzero_P$, iterating
\cref{eq:sm-iterate-error} and using \cref{eq:sm-objective-lipschitz} gives
\begin{equation*}
    |J_\lambda(\vv^{(L)})-J_\lambda(\vu^{(L)})|\leq\gamma/2.
\end{equation*}
Combining this estimate with \cref{eq:sm-ista-rate} gives the required
objective gap of at most $\gamma$ for $\vv=\vv^{(L)}$.

Finally, append a $(L+1)$-th layer consisting of the attention block from \cref{lem:sm-readout} (with readout error $\gamma$) and a zero MLP block ($\mW_1,\mW_2=\vzero$), which acts as the identity. The block writes $\widehat y_{\rm q}$ into the label coordinate of the final token, so $\widehat y_{\rm TF}=\widehat y_{\rm q}$ by \cref{eq:sm-readout-def}. This establishes \cref{eq:sm-transformer-output-error}. Moreover, for any input prompt, the clipping step ensures $\norm{\vv^{(L)}}_\infty\leq R$, and \cref{lem:sm-readout} combined with the Cauchy--Schwarz inequality yields
\begin{equation*}
|\widehat y_{\rm TF}|\leq|\vpsi(\vX_{N+1})^\top\vv^{(L)}|
\leq R\sqrt P\norm{\vpsi(\vX_{N+1})}_2,
\end{equation*}
establishing the global bounds in \cref{lem:sm-transformer-lasso-main}.
\end{proof}

\end{document}